\documentclass{article} 
\usepackage[nowritingstyle]{preprint}
\usepackage{microtype}

\definecolor{paperNavy}{HTML}{234B70}
\definecolor{paperBlue}{HTML}{326A9C}
\definecolor{paperSky}{HTML}{709DC2}
\colorlet{accent}{paperNavy}
\colorlet{linkgreen}{paperBlue}
\colorlet{linkblue}{paperBlue}
\colorlet{oursgreen}{paperBlue}
\colorlet{oursblue}{paperBlue}
\colorlet{oursteal}{paperSky}
\colorlet{gaingreen}{paperBlue}
\colorlet{gainblue}{paperBlue}
\definecolor{oursbg}{HTML}{E8F0F8}
\definecolor{abstractbg}{HTML}{EFF4FA}

\usepackage{amsmath,amsfonts,bm}

\def\eqref#1{equation~\ref{#1}}

\def\1{\bm{1}}

\DeclareMathAlphabet{\mathsfit}{\encodingdefault}{\sfdefault}{m}{sl}
\SetMathAlphabet{\mathsfit}{bold}{\encodingdefault}{\sfdefault}{bx}{n}

\usepackage{hyperref}
\usepackage{url}
\usepackage{graphicx}
\usepackage{booktabs}
\usepackage[table]{xcolor}
\usepackage{wrapfig}
\usepackage{pifont}
\definecolor{lightblue}{RGB}{225,240,250}
\definecolor{bestblue}{RGB}{224,238,250}   
\definecolor{secondblue}{RGB}{224,247,244} 

\definecolor{myblue}{RGB}{18,79,158}
\hypersetup{hidelinks,colorlinks=true, citecolor=myblue,}

\definecolor{darkgreen}{RGB}{0,110,60}
\definecolor{lightblue}{RGB}{226,238,248}

\usepackage{etoc}
\usepackage{amsmath,amssymb,amsthm}
\newtheorem{proposition}{Proposition}[section]

\theoremstyle{definition}

\usepackage{longtable}
\usepackage{array}
\usepackage{algorithm}
\usepackage{algpseudocode}

\usepackage{multirow}
\usepackage{colortbl}
\usepackage{makecell}
\usepackage{arydshln}
\usepackage{subcaption}
\usepackage{caption}

\usepackage{siunitx}
\definecolor{steel}{HTML}{2E5A88}   
\definecolor{faint}{HTML}{6B7683}   
\newcommand{\gpos}[1]{\textcolor{steel}{$+#1$}}

\definecolor{green(pigment)}{rgb}{0.0, 0.65, 0.31}
\definecolor{red(pigment)}{rgb}{0.96, 0.59, 0.48}

\newcommand{\xmark}{\textcolor{red(pigment)}{\ding{55}}}
\newcommand{\cmark}{\textcolor{green(pigment)}{\ding{51}}}

\usepackage[most]{tcolorbox}
\usepackage{enumitem}
\usepackage{fontawesome5}          

\definecolor{boxbg}{HTML}{F7F9FC}    
\definecolor{boxedge}{HTML}{C7D4E2}  
\definecolor{chipbg}{HTML}{E8EEF6}   

\definecolor{titlebg}{HTML}{E9EFF6}  

\newtcolorbox{promptbox}[1]{
  enhanced, breakable,
  colback=boxbg, colframe=boxedge,
  boxrule=0.5pt, arc=1.6mm,
  left=8pt, right=8pt, top=6pt, bottom=7pt,
  title={\faTerminal\ \ \textsc{#1}},
  coltitle=steel, colbacktitle=titlebg,   
  fonttitle=\footnotesize\bfseries,
  toptitle=3.5pt, bottomtitle=3.5pt, lefttitle=8pt,
  before skip=8pt, after skip=8pt
}
\newcommand{\ptag}[1]{\textcolor{steel}{\textbf{\footnotesize #1}}}
\newcommand{\aid}[1]{\colorbox{chipbg}{\textcolor{steel}{\footnotesize\texttt{\textbf{#1}}}}}
\newcommand{\aname}[1]{\texttt{\footnotesize #1}}

\title{Focused Omni-Modal GUI Agents}

\author[1]{Fuyao Zhang}
\author[1]{Wenwen He}
\author[1]{Jiaming Zhang}
\author[1]{Yurong Hao}
\author[1]{Xuan Wang}
\author[1]{Guowei Guan}
\author[1]{Che Wang}
\author[2]{Longtao Huang}
\author[1]{Wei Yang Bryan Lim}
\affiliation[1]{Nanyang Technological University}
\affiliation[2]{Alibaba Group}
\runninghead{Focused Omni-Modal GUI Agents}

\newcommand{\projectgithub}{https://github.com/fyzhang1/FocusGUI}
\newcommand{\projecthomepage}{https://fyzhang1.github.io/FocusGUI}

\logos{%
  \logoimage[0.34in]{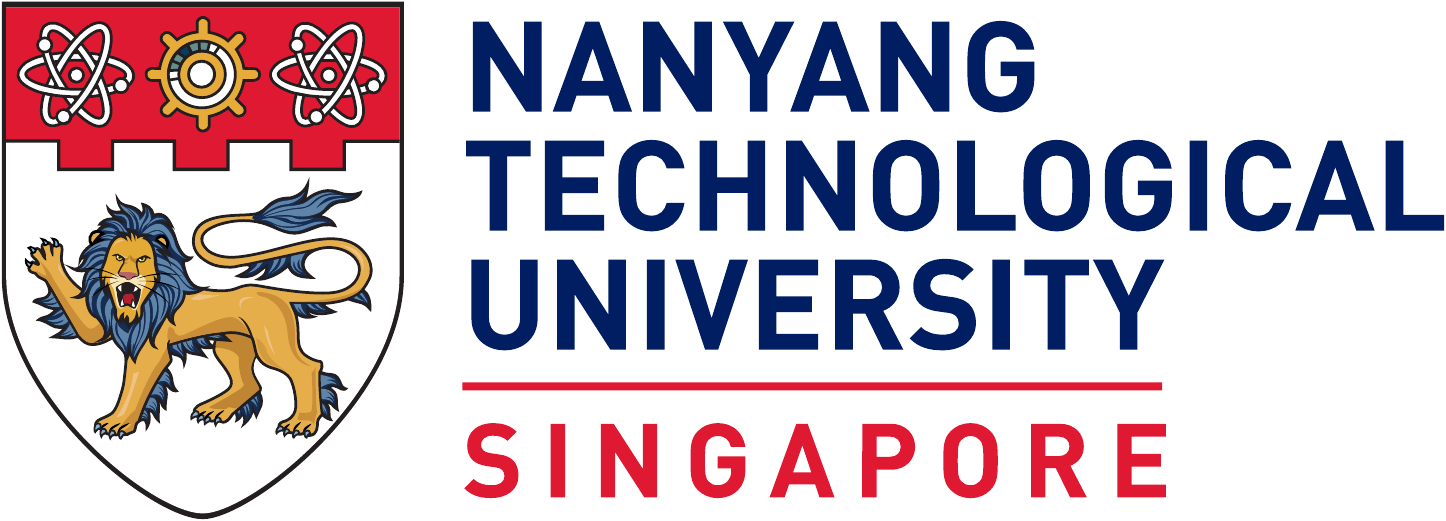}%
  \hspace{1.5em}%
  \logoimage[0.34in]{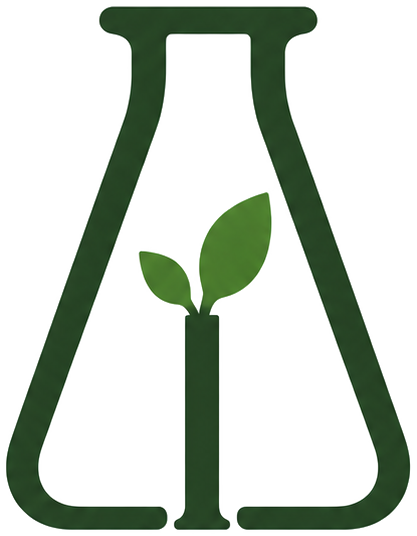}\hspace{0.45em}%
  \logoimage[0.105in]{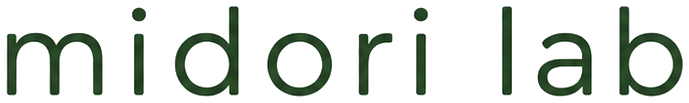}%
  \hspace{1.5em}%
  \logoimage[0.16in]{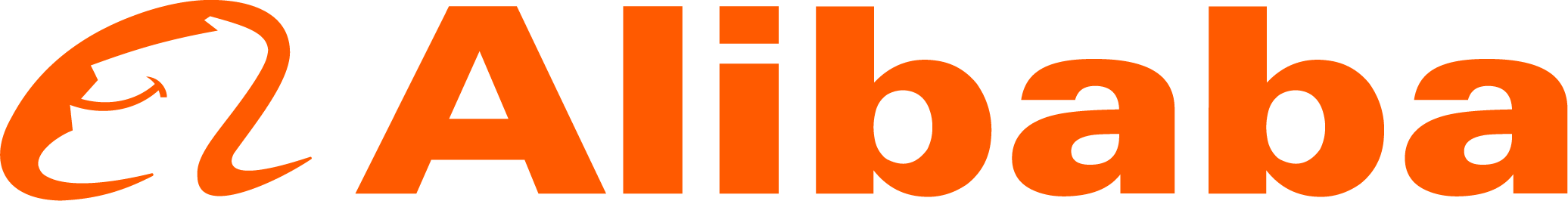}}

\authornote{{\sffamily\hypersetup{urlcolor=black}%
  \href{\projectgithub}{\textcolor{black}{\faGithub\hspace{0.4em}GitHub}}%
  \hspace{1.6em}%
  \href{\projecthomepage}{\textcolor{black}{\faGlobe\hspace{0.4em}Homepage}}}}

\AtBeginDocument{%
  \renewtcolorbox{preprintabstractbox}{%
    enhanced,colback=white,colframe=paperNavy,
    boxrule=1.1pt,arc=2.5mm,left=6mm,right=6mm,top=4mm,bottom=4mm,
    grow to left by=1.5pt,grow to right by=1.5pt,
    before skip=2mm,after skip=6mm}}

\begin{document}
\etocdepthtag.toc{mainmatter}

\maketitle

\begin{abstract}
Graphical User Interface (GUI) agents increasingly act over streaming omni-modal observations spanning screenshots, audio, video, interaction history, and device states. However, exposing this complete context uniformly at every step ignores that relevance is transient and sparse: the signal determining one action can become clutter at the next. Because executed actions feed back into the context, these errors compound into Sequential Distraction, driving the agent into irreversible trajectory drift.  We formulate the underlying requirement as Focused Perception: a representation that is step-adaptive, action-sufficient, and non-intrusive. We realize this in \textbf{FocusGUI}, a lightweight adapter that dynamically prioritizes multimodal channels against a step-specific decision reference. Instead of uniformly processing all inputs, it compresses the most relevant evidence into a compact latent and subtly injects it into the frozen decoder's embeddings, guiding the model's attention without altering its native parameter space. FocusGUI trains from action supervision alone, requiring neither modality-relevance labels nor backbone fine-tuning. Across four omni-modal backbones, it raises overall Type Match by $+33.2$ and Exact Match by $3.0\times$ over the frozen backbone on average, and Type Match by $+30.1$ over the strongest existing enhancement method.
\end{abstract}

\begin{figure*}[ht]
\centering
\includegraphics[width=1.0\linewidth]{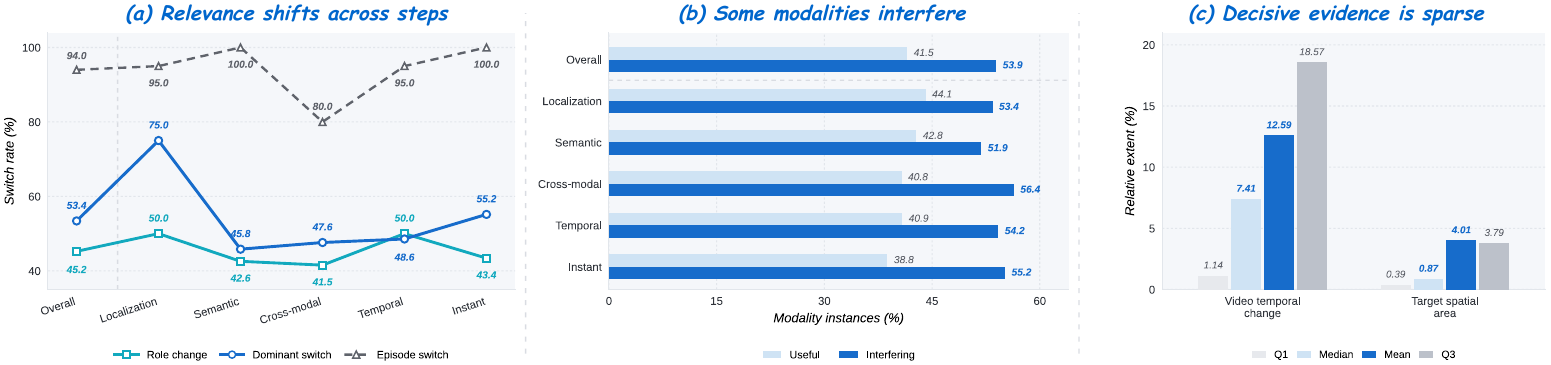}
\caption{\textbf{Relevance is transient across steps and sparse within them.}
Across multi-step episodes, we systematically ablate individual modalities from the input of Qwen2.5-Omni and record the resulting change in the log-likelihood of the ground-truth action. A modality is \emph{useful} if its ablation drops this likelihood, and \emph{interfering} if its ablation increases it; the \emph{dominant} modality is the one whose absence causes the steepest drop. 
(a) The dominant modality switches across 45.3\% of adjacent steps, with 94.0\% of episodes containing at least one switch. 
(b) 90.1\% of steps contain at least one interfering modality. 
(c) Based on annotations, the target element occupies a median of a mere 0.87\% of the screen area, and video content changes by a median of just 7.41\% within a step.}
\label{fig:motivation}
\vspace{-1em}
\end{figure*}

\section{Introduction}

Graphical User Interface (GUI) agents are shifting from parsing static screenshots to acting on dynamic, streaming observations across multiple modalities~\citep{henry2026omniguibenchmarkingguiagents, jang2025videowebarena, yao2026benchmarkinglivingscreennativeguiagents}. In realistic computing environments, the evidence needed for the next action need not be confined to the current visual frame. An unexpected notification chime might demand an immediate context switch, a specific transition in a live video can dictate when to click, and spoken instructions must be continuously mapped to visual changes on the screen~\citep{bytedance2026seedrealtime, li2026weavebenchlonghorizonrealworldbenchmark, yuan2026osworld20benchmarkingcomputer, chen2025gui}. Contemporary omni-modal models can ingest screenshots, video streams, audio, interaction history, and device states within a single network~\citep{xu2025qwen25omnitechnicalreport, fu2025vita15gpt4olevelrealtime, cui2026minicpmo45realtimefullduplex}. This has naturally led to a default paradigm: exposing the complete multimodal observation to the model at every interaction step and leaving it to implicitly determine what matters while generating the action.

Leaving this selection entirely to the decoder assumes that it can reliably rediscover what matters from the complete observation at every interaction step. Our analysis reveals why this is difficult (Fig.~\ref{fig:motivation}). When individual modalities are removed during multi-step interaction, the dominant modality switches across nearly half of adjacent steps; even within a single step, only a small fraction of the observation may support the decision, while the rest can become distracting. Relevance is therefore fundamentally a property of the \emph{step}, not a static property of the observation. Yet the decoder is forced to rediscover this relevance at every step, without any explicit signal of what has changed. In a closed-loop agent, such misreads do not end with the current prediction: the executed action changes the environment and feeds back into the observation and history of the next step, allowing errors to compound over time. We refer to this compounding failure as \textbf{Sequential Distraction}.

The most intuitive remedy is to perform \textit{subtraction} on the inputs. This is exactly what existing methods attempt: cropping parts of the current screen to retain only candidate regions of interest~\citep{lu2024omniparserpurevisionbased, Luo_2025_ICCV, Liu_2026_CVPR, cheng2024seeclick}. However, this strategy of giving less fundamentally misaligns with the nature of streaming interaction in two ways. First, the decisive signal could emerge from audio, video, or device status at any moment, channels entirely outside the scope of spatial cropping. Second, even when the signal resides on the screen, the crudely cropped background often contains the essential contextual evidence needed to ground accurate coordinates and text parameters. The problem is not that the decoder is given too much, but that it lacks guidance on how to prioritize it. A proper focus mechanism should change how the decoder \emph{reads} the observation, rather than altering what it \emph{is given}.

\begin{figure*}
    \centering
    \includegraphics[width=0.95\linewidth]{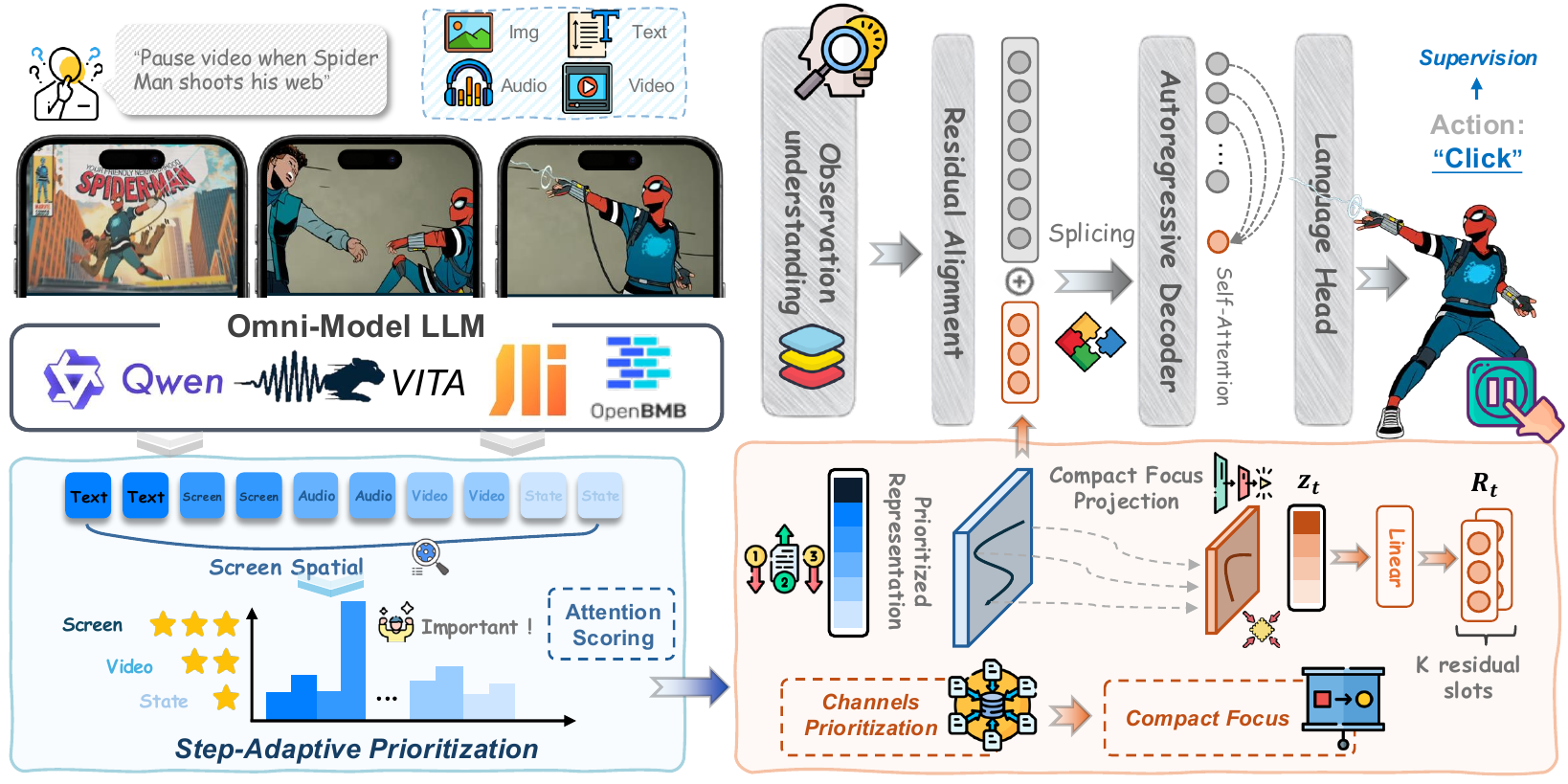}
    \caption{\textbf{FocusGUI.} The frozen backbone's original pathway is left intact (top). From the same observation, FocusGUI exposes perceptual channels, scores them against a decision reference recomputed at every step, compresses the result into a compact focus latent, and adds it as a zero-initialized
residual at the $K$ positions preceding the action boundary.}
    \label{fig:overview}
\end{figure*}

We realize this principle in \textbf{FocusGUI} (Fig.~\ref{fig:overview}). Since the decoder can already read the full observation and what it lacks is a signal of what matters at the current step, FocusGUI leaves both the backbone and its input untouched and supplies that signal alongside them. At every step, it exposes the observation as separable perceptual channels, scores each against a reference derived from the current observation as a whole, and compresses the prioritized channels into a compact focus latent. The latent enters as a zero-initialized residual at the boundary where the decoder begins to generate the action. This residual reallocates attention over the entire observation without altering it, guiding both the choice of action and the grounding of its arguments. Training uses next-action supervision alone, with no labels indicating which modality matters.

On average across four omni-modal backbones, FocusGUI raises Type Match from 31.0\% to 64.2\% and Exact Match from 11.2\% to 33.5\% over the frozen backbone, and reaches 1.9$\times$ the Type Match of the strongest enhancement method. The gains grow with interaction depth, as Sequential Distraction predicts: from the first step to the fourth, the frozen Qwen2.5-Omni loses half of its Type Match while FocusGUI retains 86.8\%. Our contributions are as follows:

\begin{itemize}
    \item An empirical account of \textbf{Sequential Distraction}: in multi-step omni-modal GUI interaction, relevance is a step-level property, transient across steps and sparse within them, and because actions re-enter the observation, per-step misreads compound into trajectory drift.
    \item A distinction between the evidence an agent is given and the way its decoder reads it, showing why spatial refinement cannot satisfy both the selection of an action and the grounding of its arguments, and deriving from this the form a solution must take.
    \item \textbf{FocusGUI}, a zero-initialized residual adapter at the action boundary of a frozen backbone, trained from action supervision alone, whose gains across four backbones and five task categories are distributed as the account predicts.
\end{itemize}

\section{Method}
\label{sec:method}

\paragraph{Problem Formulation and Context-Preserving Focus}
\label{sec:problem_formulation} At the interaction step $t$, the agent receives an omni-modal observation
$O_t=\{g,I_t,V_t,A_t,H_t,S_t\}$, consisting of the user goal $g$, current interface $I_t$, temporal visual observation $V_t$, acoustic observation $A_t$, interaction history $H_t$, and device state $S_t$. It predicts an executable action $y_t=(y_{t,1},\ldots,y_{t,L_t})$, represented as a token sequence specifying both the action primitive and its arguments. Let $F_{\Theta_0}$ denote a pretrained omni-modal backbone whose parameters
$\Theta_0$ remain frozen, and let $E_t \in \mathbb{R}^{N_t \times d_{\mathrm{model}}}$ denote the embedding sequence presented to its decoder. Section~1 committed to realizing focus in how the decoder reads rather than on what it is shown, and that commitment fixes the form of the mechanism. FocusGUI neither replaces nor prunes any part of $O_t$. From the same observation, it constructs a compact focus representation $z_t = f_\phi(O_t)$, maps it to a residual $R_t = A_\phi(z_t)$, and adds that residual to a small set of existing embedding positions. The decoder therefore operates on
\begin{equation}
    \widetilde{E}_t
    =
    E_t+\mathcal{S}_{J_t}(\gamma R_t),
    \qquad
    p_{\Theta_0,\phi}(y_t\mid O_t)
    =
    p_{\Theta_0}(y_t\mid \widetilde{E}_t),
    \label{eq:overview}
\end{equation}
where $\mathcal{S}_{J_t}(\cdot)$ places the residual at positions $J_t$ and is zero elsewhere. This separates \emph{observation preservation} from \emph{decision guidance}.
The original pathway retains the complete multimodal context, while the FocusGUI pathway learns how that context should be emphasized for the next action. Two properties of $z_t$ follow from this division of labor. It is not
required to reconstruct the observation or to encode any execution detail,
since every pixel, coordinate, and transcript remains available to the decoder
through the unmodified pathway. It can therefore be made small, because its
role is to condition a read rather than to carry evidence. Appendix~\ref{app:theory} shows that such a residual selects the primitive by
changing what the decoder reads, but corrects arguments only by reallocating
attention over evidence positions Eq.~(1) leaves untouched.

\paragraph{Step-Adaptive Evidence Organization}

A priority cannot be defined inside a fused representation, so the observation
must first be exposed as separable evidence. We write this as perceptual
channels $X_t = \{x_t^{(m)}\}_{m \in \mathcal{M}}$, where $\mathcal{M}$ indexes
the sources' FocusGUI scores, including the backbone's integrated hidden state
and a spatial screen channel alongside the individual modalities.
Each channel is mapped into a shared $d$-dimensional space through a source-specific projector,
$e_t^{(m)}=P_m(x_t^{(m)})\in\mathbb{R}^{d}$.
The separate projectors accommodate the native statistics of different sources while making their representations comparable. The usefulness of a channel, however, depends not only on what it contains but also on the surrounding task state. We therefore derive a decision reference $q_t=P_q(e_t^{(\mathrm{all})})$ from the integrated representation of the complete observation. Here, $e_t^{(\mathrm{all})}$ provides a \emph{joint semantic context}: it summarizes the current observation against which source-specific evidence is evaluated. For each channel $m$, FocusGUI computes its compatibility with this joint context and normalizes the resulting scores into step-specific priorities:
\begin{equation}
\begin{aligned}
    s_t^{(m)}
    &=
    v^\top
    \tanh\!\left(
        W_e(e_t^{(m)}+r_m)+W_q q_t
    \right), \\
    \alpha_t^{(m)}
    &=
    \frac{\exp(s_t^{(m)}/\tau)}
    {\sum_{j\in\mathcal{M}}\exp(s_t^{(j)}/\tau)},
    \qquad
    c_t
    =
    \sum_{m\in\mathcal{M}}
    \alpha_t^{(m)}e_t^{(m)},
\end{aligned}
\label{eq:prioritization}
\end{equation}
where $r_m$ is a learned channel-identity embedding and $\tau$ is the softmax temperature. Because both the perceptual channels and the joint semantic context are recomputed at every interaction step, the relative emphasis among information sources can change as the task evolves. Nothing in Eq.~(2) carries state across steps: drift is produced one step at a
time, so removing the corruption at each step removes its accumulation. The priorities $\alpha_t^{(m)}$ should be interpreted as how FocusGUI organizes evidence within its auxiliary pathway, rather than as a complete explanation of how the frozen backbone internally uses each modality.

\paragraph{Compact Focus and Residual Alignment}
\label{sec:compact_residual}
The prioritized representation $c_t$ determines how different evidence sources contribute to the auxiliary pathway, but still lies in the full shared channel space.
We therefore compress it into a lower-dimensional focus representation: $z_t=P_s(c_t)\in\mathbb{R}^{d_z}, \qquad d_z<d.$ This projection constrains the capacity of the \emph{auxiliary conditioning signal}, rather than the information accessible to the backbone.
Because the complete observation remains available through the original multimodal pathway, $z_t$ need not independently encode every visual detail, coordinate, or text argument required for execution. Instead, it provides a compact summary of what should receive greater emphasis for the current action. To make this focus representation directly usable by the frozen decoder, we align it with the decoder's native embedding space at the action boundary.
Let $b_t$ denote the position at which action generation begins, and let $J_t=\{b_t-K,\ldots,b_t-1\}$ denote the $K$ preceding embedding positions.
FocusGUI expands $z_t$ into a residual over this window and adds it to the existing embeddings:
\begin{equation}
    R_t
    =
    \operatorname{Reshape}(W_p z_t+b_p)
    \in\mathbb{R}^{K\times d_{\mathrm{model}}},
    \qquad
    \widetilde{E}_{t,J_t}
    =
    E_{t,J_t}+\gamma R_t.
    \label{eq:residual_alignment}
\end{equation}

Injecting the residual immediately before action generation makes the focus signal available to all subsequent action tokens without introducing a separate prediction head or modifying the output vocabulary.
The additive form is equally important: it preserves the original embeddings and uses the focus representation only as an additional conditioning signal. We initialize $W_p$ and $b_p$ to zero, such that $R_t=0$ and $\widetilde{E}_t=E_t$ at the beginning of training.
FocusGUI therefore starts from the original frozen-model input and gradually learns observation-conditioned deviations through optimization. Importantly, $J_t$ consists of existing prompt positions: FocusGUI neither appends textual hints nor removes any multimodal input.

\paragraph{Learning Focus from Action Supervision}
\label{sec:learning_focus}

FocusGUI requires no annotation specifying which modality or perceptual channel should be emphasized.
Instead, all components are learned jointly from supervision on the next executable action.
Let $\phi$ denote the trainable parameters of the channel projectors, step-adaptive prioritization, compact projection, and residual alignment modules.
We optimize the average action-token negative log-likelihood
\begin{equation}
    \mathcal{L}_{\mathrm{act}}(\phi)
    =
    -\frac{1}{|\mathcal{D}_{\mathrm{tr}}|}
    \sum_{t\in\mathcal{D}_{\mathrm{tr}}}
    \frac{1}{L_t}
    \sum_{k=1}^{L_t}
    \log
    p_{\Theta_0,\phi}
    \left(
        y_{t,k}
        \mid
        y_{t,<k},O_t
    \right).
    \label{eq:objective}
\end{equation}

The loss is evaluated only on action tokens.
Gradients propagate through the frozen decoder to $\phi$, while the omni-modal backbone, decoder, and language-model head receive no parameter updates.
Thus, a single action-generation objective jointly learns
\emph{which evidence to emphasize},
\emph{how to summarize it},
and \emph{how to express that summary in the decoder's native representation space}. At inference time, FocusGUI constructs the perceptual channels from the current observation, evaluates them under the current joint semantic context, forms the compact focus representation, and generates the corresponding residual.
The frozen decoder then autoregressively produces the next executable action through its original generation pathway.

\section{Experiments}
\subsection{Experimental Setup}

\paragraph{Dataset and Metrics.}
We evaluate on OmniGUI~\citep{henry2026omniguibenchmarkingguiagents}, a benchmark of omni-modal smartphone GUI interactions across 29 applications; we use 708 episodes with 2{,}572 action steps. Each step pairs a screenshot with synchronous audio, video, and action history, so the decisive signal can come from any of the four streams. We report Type Match (TM), Exact Match (EM), and Goal Progress (GP). TM checks whether the predicted action primitive matches the ground truth; EM further requires predicted coordinates to fall within the target element's bounding box and generated text to match the reference. GP is the fraction of correctly executed steps per episode, averaged across episodes.

\begin{table*}[t]
\centering
\small
\renewcommand{\arraystretch}{1.42}
\setlength{\tabcolsep}{1.15mm}

\caption{
Main results across omni-modal backbones and GUI task categories.
FocusGUI rows are highlighted, and the best score in each column
within each backbone block is bolded.
}
\label{tab:main_results}

\setlength{\tabcolsep}{2.2pt}
\scriptsize

\resizebox{\textwidth}{!}{%
\begin{tabular}{l||ccc|ccc:ccc:ccc:ccc:ccc}
\Xhline{1.2pt}

\rowcolor{gray!30}
& \multicolumn{3}{c|}{Overall}
& \multicolumn{3}{c:}{Localization}
& \multicolumn{3}{c:}{Semantic}
& \multicolumn{3}{c:}{Cross-modal}
& \multicolumn{3}{c:}{Temporal}
& \multicolumn{3}{c}{Instant} \\

\rowcolor{gray!30}
Methods
& TM$\uparrow$ & EM$\uparrow$ & GP$\uparrow$
& TM$\uparrow$ & EM$\uparrow$ & GP$\uparrow$
& TM$\uparrow$ & EM$\uparrow$ & GP$\uparrow$
& TM$\uparrow$ & EM$\uparrow$ & GP$\uparrow$
& TM$\uparrow$ & EM$\uparrow$ & GP$\uparrow$
& TM$\uparrow$ & EM$\uparrow$ & GP$\uparrow$ \\

\Xhline{0.8pt}

Qwen2.5-Omni
& 49.61 & 12.60 & 5.50
& 57.32 & 12.20 & 4.94
& 48.65 & 17.12 & 6.75
& 42.20 & 9.17 & 3.17
& 45.97 & 12.10 & 7.40
& 58.54 & 12.20 & 5.07 \\

+DRS-GUI
& 49.41 & 6.69 & 2.97
& 51.22 & 6.10 & 3.63
& 50.45 & 5.41 & 1.23
& 45.87 & 5.50 & 2.79
& 49.19 & 9.68 & 4.95
& 51.22 & 6.10 & 1.87 \\

+OmniParser
& 48.43 & 7.68 & 3.87
& 51.22 & 7.32 & 4.51
& 49.55 & 7.21 & 1.87
& 44.95 & 5.50 & 2.41
& 46.77 & 10.48 & 5.83
& 51.22 & 7.32 & 4.48 \\

+RegionFocus
& 48.23 & 7.68 & 4.07
& 52.44 & 4.88 & 3.82
& 49.55 & 7.21 & 2.65
& 44.04 & 6.42 & 3.46
& 46.77 & 11.29 & 5.21
& 50.00 & 7.32 & 5.13 \\

+GUI-Actor
& 50.20 & 13.78 & 5.54
& 58.54 & 13.41 & 5.18
& 47.75 & 18.02 & 6.36
& 43.12 & 10.09 & 3.63
& 46.77 & 12.90 & 7.08
& 59.76 & 14.63 & 5.47 \\

\rowcolor{bestblue}
+FocusGUI
& \textbf{71.65} & \textbf{40.16} & \textbf{17.49}
& \textbf{78.05} & \textbf{45.12} & \textbf{30.06}
& \textbf{68.47} & \textbf{39.64} & \textbf{14.24}
& \textbf{67.89} & \textbf{33.03} & \textbf{7.47}
& \textbf{76.61} & \textbf{46.77} & \textbf{26.35}
& \textbf{67.07} & \textbf{35.37} & \textbf{7.97} \\

\hline

VITA-1.5
& 13.98 & 9.65 & 6.09
& 12.20 & 7.32 & \textbf{7.04}
& 19.82 & 13.51 & 6.59
& 11.01 & 6.42 & 2.69
& 12.90 & 9.68 & 8.85
& 13.41 & 10.98 & \textbf{5.07} \\

+DRS-GUI
& 10.43 & 6.30 & 3.14
& 8.54 & 3.66 & 1.26
& 9.01 & 6.31 & 3.05
& 10.09 & 3.67 & 1.76
& 12.90 & 10.48 & 5.58
& 10.98 & 6.10 & 3.89 \\

+OmniParser
& 17.13 & 7.48 & 3.82
& 18.29 & 3.66 & 1.26
& 17.12 & 7.21 & 3.77
& 14.68 & 4.59 & 2.93
& 17.74 & 12.10 & 6.40
& 18.29 & 8.54 & 4.59 \\

+RegionFocus
& 12.20 & 7.68 & 3.69
& 9.76 & 4.88 & 1.44
& 10.81 & 7.21 & 3.63
& 10.09 & 4.59 & 2.75
& 15.32 & 12.10 & 5.77
& 14.63 & 8.54 & 4.78 \\

+GUI-Actor
& 14.57 & 11.02 & 5.39
& 17.07 & 9.76 & 1.23
& 22.52 & 18.02 & \textbf{11.14}
& 9.17 & 6.42 & \textbf{3.82}
& 12.90 & 11.29 & 5.94
& 10.98 & 8.54 & 4.35 \\

\rowcolor{bestblue}
+FocusGUI
& \textbf{58.66} & \textbf{26.38} & \textbf{6.84}
& \textbf{56.10} & \textbf{18.29} & 6.79
& \textbf{54.05} & \textbf{28.83} & 8.76
& \textbf{55.96} & \textbf{24.77} & 3.71
& \textbf{64.52} & \textbf{30.65} & \textbf{9.69}
& \textbf{62.20} & \textbf{26.83} & 4.71 \\

\hline

MiniCPM-o-4.5
& 36.02 & 13.58 & 7.55
& 35.37 & 13.41 & 9.97
& 35.14 & 9.91 & 5.15
& 35.78 & 15.60 & 3.98
& 34.68 & 12.90 & 11.04
& 40.24 & 17.07 & 7.68 \\

+DRS-GUI
& 36.42 & 12.60 & 5.47
& 37.80 & 10.98 & 3.40
& 36.94 & 14.41 & 5.24
& 31.19 & 8.26 & 4.41
& 37.90 & 16.94 & 8.62
& 39.02 & 10.98 & 5.37 \\

+OmniParser
& 37.20 & 12.40 & 6.43
& 41.46 & 9.76 & 4.84
& 36.04 & 12.61 & 5.84
& 30.28 & 8.26 & 5.02
& 38.71 & 16.94 & 9.26
& 41.46 & 13.41 & 6.96 \\

+RegionFocus
& 37.01 & 13.19 & 6.69
& 40.24 & 10.98 & 5.62
& 36.94 & 15.32 & 7.27
& 30.28 & 9.17 & 5.32
& 39.52 & 17.74 & 9.32
& 39.02 & 10.98 & 5.57 \\

+GUI-Actor
& 36.61 & 14.17 & 7.64
& 36.59 & 14.63 & 9.47
& 34.23 & 10.81 & 5.63
& 36.70 & 16.51 & 4.23
& 35.48 & 12.10 & 10.84
& 41.46 & 18.29 & 8.03 \\

\rowcolor{bestblue}
+FocusGUI
& \textbf{67.72} & \textbf{41.54} & \textbf{16.76}
& \textbf{81.71} & \textbf{47.56} & \textbf{21.73}
& \textbf{68.47} & \textbf{45.05} & \textbf{19.94}
& \textbf{60.55} & \textbf{34.86} & \textbf{9.15}
& \textbf{67.74} & \textbf{41.13} & \textbf{18.68}
& \textbf{62.20} & \textbf{40.24} & \textbf{14.49} \\

\hline

Baichuan-Omni-1.5
& 24.41 & 8.86 & 4.32
& 26.83 & 4.88 & 0.00
& 16.22 & 8.11 & 1.72
& 22.02 & 11.93 & 6.77
& 28.23 & 10.48 & 7.66
& 30.49 & 7.32 & 4.71 \\

+DRS-GUI
& 27.76 & 5.12 & 2.26
& 32.93 & 3.66 & 2.76
& 26.13 & 5.41 & 2.08
& 21.10 & 2.75 & 1.19
& 29.03 & 8.06 & 2.89
& 31.71 & 4.88 & 2.29 \\

+OmniParser
& 32.09 & 6.10 & 2.21
& 37.80 & 4.88 & 3.01
& 31.53 & 6.31 & 0.78
& 26.61 & 5.50 & 1.82
& 30.65 & 8.06 & 3.26
& 36.59 & 4.88 & 1.96 \\

+RegionFocus
& 24.61 & 5.51 & 2.47
& 28.05 & 3.66 & 2.41
& 22.52 & 5.41 & 1.57
& 19.27 & 3.67 & 0.87
& 25.81 & 8.06 & 4.45
& 29.27 & 6.10 & 2.84 \\

+GUI-Actor
& 31.10 & 11.02 & 5.35
& 34.15 & 8.54 & 2.91
& 24.32 & 10.81 & 3.11
& 28.44 & 12.84 & \textbf{6.92}
& 33.06 & 12.10 & 8.57
& 37.80 & 9.76 & \textbf{5.24} \\

\rowcolor{bestblue}
+FocusGUI
& \textbf{58.86} & \textbf{25.98} & \textbf{8.73}
& \textbf{56.10} & \textbf{18.29} & \textbf{3.70}
& \textbf{58.56} & \textbf{33.33} & \textbf{15.27}
& \textbf{55.05} & \textbf{21.10} & 4.09
& \textbf{62.10} & \textbf{31.45} & \textbf{15.21}
& \textbf{62.20} & \textbf{21.95} & 3.62 \\

\Xhline{1.2pt}
\end{tabular}%
}

\end{table*}

\paragraph{Models.}
We evaluate four representative open-source omni-modal backbones: Qwen2.5-Omni~\citep{xu2025qwen25omnitechnicalreport}, VITA-1.5~\citep{fu2025vita15gpt4olevelrealtime}, MiniCPM-o~4.5~\citep{cui2026minicpmo45realtimefullduplex}, and Baichuan-Omni-1.5~\citep{li2025baichuan}. The four backbones support textual, visual, auditory, and temporal inputs but differ in architecture and multimodal representation, which lets the comparison probe the generality of FocusGUI across backbone families. For each backbone, the base model stays frozen, and training updates only the lightweight FocusGUI head and action predictors, using the same episode-disjoint split across all four models.

\paragraph{Baselines.} We compare against four methods, each applied to all
four backbones without modifying their weights. DRS-GUI~\citep{Liu_2026_CVPR},
OmniParser~\citep{lu2024omniparserpurevisionbased}, and
RegionFocus~\citep{Luo_2025_ICCV} refine the visual input at inference by
searching over candidate regions, parsing the screenshot into structured
elements, and zooming into attended regions.
GUI-Actor~\citep{wu2025guiactorcoordinatefreevisualgrounding} instead learns an
action head over the backbone's visual tokens, trained on the same annotations
and split as FocusGUI with the backbone frozen. All methods receive the same
observation and decoding procedure, differing only in how that observation is
organized before the action is generated. The trained head separates two
explanations for any gain: trainable parameters on this data, which GUI-Actor
also has, and step-adaptive organization, which FocusGUI adds.

\paragraph{Implementation Details.} Evidence channels are projected into a shared 128-dimensional space and weighted by the full-modal query with softmax temperature $\tau=1.0$; the weighted representation is compressed to a 64-dimensional action-sufficient embedding and expanded into $K=4$ zero-initialized residual tokens added to the last four prompt embeddings before the action boundary, introducing 1.15M--1.45M trainable parameters depending on the backbone. With the encoder, decoder, and LM head frozen, we train only these modules for three epochs with AdamW under the teacher-forced action-token negative log-likelihood, selecting the learning rate $2\times10^{-4}$ and weight decay $10^{-3}$. At inference, the frozen decoder generates actions autoregressively with greedy decoding. Additional implementation details are provided in the Appendix~\ref{app:impl}.

\subsection{Main Results}

\textbf{Consistent gains across backbones.}
Table~\ref{tab:main_results} evaluates four omni-modal backbones under identical frozen weights. Averaged across them, FocusGUI adds 33.2 TM, 22.3 EM, and 6.6 GP points over direct inference and 30.1, 23.6, and 8.0 points over the strongest baseline at each backbone. It holds the best TM and the best EM in all 24 backbone-category cells.
 
\textbf{Why screen refinement backfires.}
The baselines fail in a way that is diagnostic rather than incidental.
DRS-GUI, OmniParser, and RegionFocus are near-neutral on TM ($+0.9$ points on
average) yet lower EM by 28\% and GP by 35\% in relative terms, in 12 of 12
cases for each metric. Narrowing a screenshot makes the action primitive more
salient, which is what TM rewards, while discarding the surrounding context
that fixes coordinates and text, which is what EM requires. None of the three
reaches audio, video, or device state, so a decision resting on those channels
stays unrecoverable however the screen is processed. This is what our framing
predicts: once relevance is a property of a step rather than of a region,
refining the region cannot restore it and may cost more than it returns.
 
\textbf{Where the gains come from.}
The structure of the gains indicates what the representation supplies. Per-category TM gains run from 37.3 points on Temporal, where the decisive signal persists across frames and a single screenshot serves worst, to 27.8 on Instant, where the decision is reactive and largely present on the current screen; EM follows the same order. Across backbones, the overall TM spread narrows from 35.6 points to 13.0, indicating that choosing the correct primitive becomes substantially less backbone-dependent under a better-organized observation. EM variation instead reflects backbone-specific argument-generation capacity. At episode level, improvements in per-step prediction translate into Goal Progress when they are sustained across the trajectory. Episode progress is gated by the weakest step in a trajectory.

\subsection{Ablation Study}

\begin{table*}[t]
\centering
\small
\renewcommand{\arraystretch}{1.52}
\setlength{\tabcolsep}{1.15mm}

\caption{
Component ablation on Qwen2.5-Omni.
CD: channel decomposition.
SP: step-adaptive prioritization.
AP: action-sufficient projection.
RA: residual alignment.
Each task category reports Type Match (TM), Exact Match (EM), and Goal Progress (GP).
}
\label{tab:ablation_prioritization}

\resizebox{\textwidth}{!}{%
\begin{tabular}{cccc||ccc|ccc:ccc:ccc:ccc:ccc}

\Xhline{1.2pt}

\rowcolor{gray!30}
\multicolumn{4}{c||}{\textbf{Components}}
& \multicolumn{3}{c|}{\textbf{Overall}}
& \multicolumn{3}{c:}{\textbf{Localization}}
& \multicolumn{3}{c:}{\textbf{Semantic}}
& \multicolumn{3}{c:}{\textbf{Cross-modal}}
& \multicolumn{3}{c:}{\textbf{Temporal}}
& \multicolumn{3}{c}{\textbf{Instant}} \\

\rowcolor{gray!30}
\textbf{CD} & \textbf{SP} & \textbf{AP} & \textbf{RA}
& \textbf{TM$\uparrow$} & \textbf{EM$\uparrow$} & \textbf{GP$\uparrow$}
& \textbf{TM$\uparrow$} & \textbf{EM$\uparrow$} & \textbf{GP$\uparrow$}
& \textbf{TM$\uparrow$} & \textbf{EM$\uparrow$} & \textbf{GP$\uparrow$}
& \textbf{TM$\uparrow$} & \textbf{EM$\uparrow$} & \textbf{GP$\uparrow$}
& \textbf{TM$\uparrow$} & \textbf{EM$\uparrow$} & \textbf{GP$\uparrow$}
& \textbf{TM$\uparrow$} & \textbf{EM$\uparrow$} & \textbf{GP$\uparrow$} \\

\hline
\hline

\xmark & \xmark & \xmark & \xmark
& 49.61 & 12.60 & 5.50
& 57.32 & 12.20 & 4.94
& 48.65 & 17.12 & 6.75
& 42.20 & 9.17 & 3.17
& 45.97 & 12.10 & 7.40
& 58.54 & 12.20 & 5.07 \\

\xmark & \xmark & \cmark & \cmark
& 66.26 & 35.00 & 12.20
& 68.49 & 40.56 & 14.55
& 66.67 & 33.64 & 11.28
& 60.55 & 30.86 & 6.58
& 72.58 & 38.58 & 20.00
& 60.29 & 31.46 & 7.13 \\

\cmark & \xmark & \cmark & \cmark
& 67.38 & 37.78 & 11.56
& 71.95 & 42.68 & 16.82
& 64.86 & 37.84 & 9.65
& 64.22 & 32.04 & 6.35
& 74.42 & 45.97 & 18.02
& 59.76 & 28.05 & 5.80 \\

\cmark & \cmark & \xmark & \cmark
& 71.14 & 38.75 & 16.11
& 78.05 & 42.68 & 25.43
& 66.67 & 37.84 & 13.84
& 67.72 & 32.70 & 6.54
& 78.23 & 47.58 & 27.50
& 65.03 & 32.93 & 7.25 \\

\cmark & \cmark & \cmark & \xmark
& 70.45 & 34.86 & 12.11
& 74.15 & 41.34 & 28.09
& 67.57 & 33.33 & 13.43
& 67.56 & 32.04 & 5.05
& 79.84 & 39.52 & 10.73
& 63.11 & 28.05 & 3.26 \\

\rowcolor{bestblue}
\cmark & \cmark & \cmark & \cmark
& \textbf{71.65} & \textbf{40.16} & \textbf{17.49}
& \textbf{78.05} & \textbf{45.12} & \textbf{30.06}
& \textbf{68.47} & \textbf{39.64} & \textbf{14.24}
& \textbf{67.89} & \textbf{33.03} & \textbf{7.47}
& \textbf{76.61} & \textbf{46.77} & \textbf{26.35}
& \textbf{67.07} & \textbf{35.37} & \textbf{7.97} \\

\Xhline{1.2pt}

\end{tabular}%
}

\end{table*}

\begin{figure*}[t]
\centering
\includegraphics[width=1.0\linewidth]{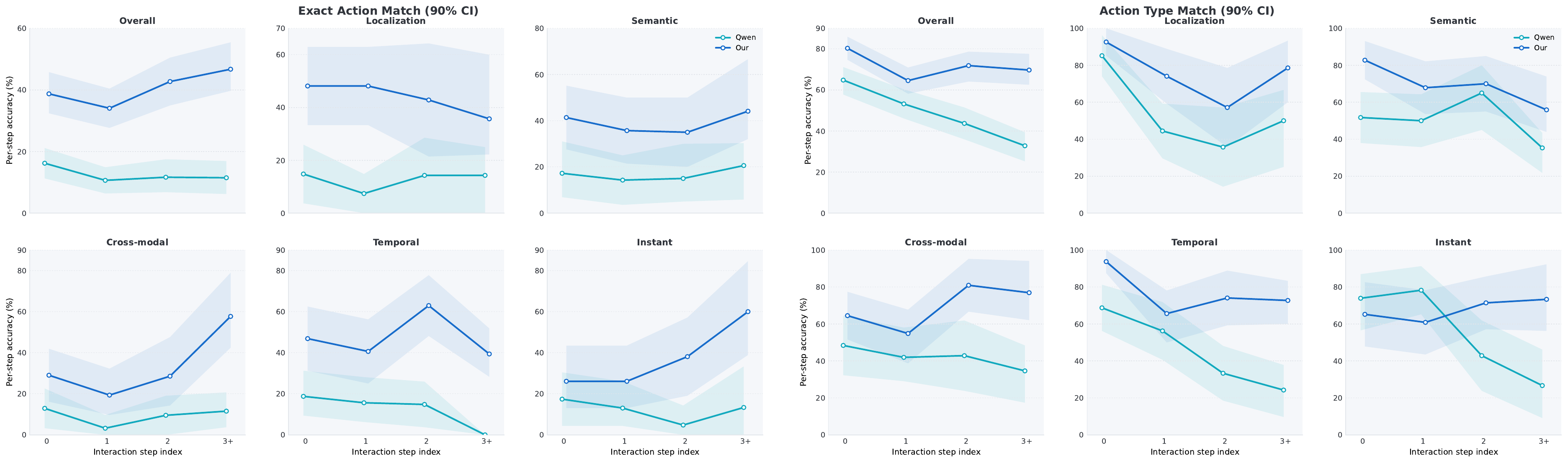}
\caption{\textbf{FocusGUI holds accuracy as trajectories lengthen.} Per-step Exact Match (left) and Type Match (right) on Qwen2.5-Omni, with steps binned by their index within the episode and $3{+}$ collecting the fourth step onward. Shaded regions are $90\%$ episode-clustered bootstrap intervals.}
\vspace{-8pt}
\label{fig:depth}
\end{figure*}

\paragraph{All four components contribute, mostly to Goal Progress.}
Table~\ref{tab:ablation_prioritization} removes each component from the full model. Overall Type Match falls by 0.51, 1.20, 4.27, and 5.39 points when action-sufficient projection, residual alignment, step-adaptive prioritization, and channel decomposition are removed, but GP falls by 1.38, 5.38, 5.93, and 5.29 points on a base roughly a quarter as large. In relative terms, every component is worth four to five times more to episode progress than to single-step action type: at most 7.5\% of TM against up to 33.9\% of GP. Their disproportionately larger effect on GP shows that the components primarily keep a trajectory from unravelling, which is what Sequential Distraction predicts.
 
\textbf{GP reveals contributions beyond TM.}
Decomposing the observation into channels without prioritizing them raises Type Match by 1.12 and Exact Match by 2.78 points over a single integrated channel. Prioritization converts this richer channel access into episode-level stability by deciding which channel to trust. Residual alignment shows the same metric asymmetry more sharply: replacing the additive residual with prompt-embedding substitution reduces EM by 5.30 and GP by 5.38 points, compared with a 1.20-point TM change. Preserving the original prompt embeddings therefore matters most for grounding the selected action primitive into correct arguments.

\subsection{Performance Across Interaction Depth}

Sequential Distraction predicts that interference accumulates. A method that
suppresses it should separate from the backbone as a trajectory lengthens,
whereas a uniformly stronger predictor would hold a constant margin. We bin
every evaluation step by its index within the episode and report per-step
accuracy with $90\%$ episode-clustered bootstrap intervals (Fig.~\ref{fig:depth}).
 
Qwen2.5-Omni falls from 64.79 TM at the first step to 32.79 beyond the third,
retaining half of its initial accuracy. FocusGUI retains 86.8\% over the same
range, and its exact-match accuracy rises rather than falls, from 38.73 to
46.72. The margin therefore widens with depth: EM gains increase monotonically
from $+22.54$ to $+35.25$, and all intervals exclude zero. This locates the
episode-level improvement of Table~\ref{tab:main_results}: Goal Progress does not rise
evenly along a trajectory but comes predominantly from steps the frozen
backbone has already lost. The effect is sharpest under temporal reasoning,
where the backbone scores 0.00 EM beyond the third step and FocusGUI reaches
39.39.
 
Instant-response tasks provide a negative control. At the first two steps, where these reactive tasks carry little accumulated history, FocusGUI is statistically indistinguishable from the backbone (-8.70 and -17.39 TM; both 90\% intervals include zero). At depths 2 and 3+, once interaction history can accumulate, FocusGUI gains $+28.57$ and $+46.67$ TM. This depth-dependent separation is the behavior expected of a mechanism targeting accumulation, rather than a uniformly stronger predictor. Restricting the analysis to episodes with at least four steps holds the episode set constant across bins, and the widening margin persists.

\begin{wrapfigure}{r}{0.52\textwidth}
    \centering
    \vspace{-8pt}
    \includegraphics[width=\linewidth]{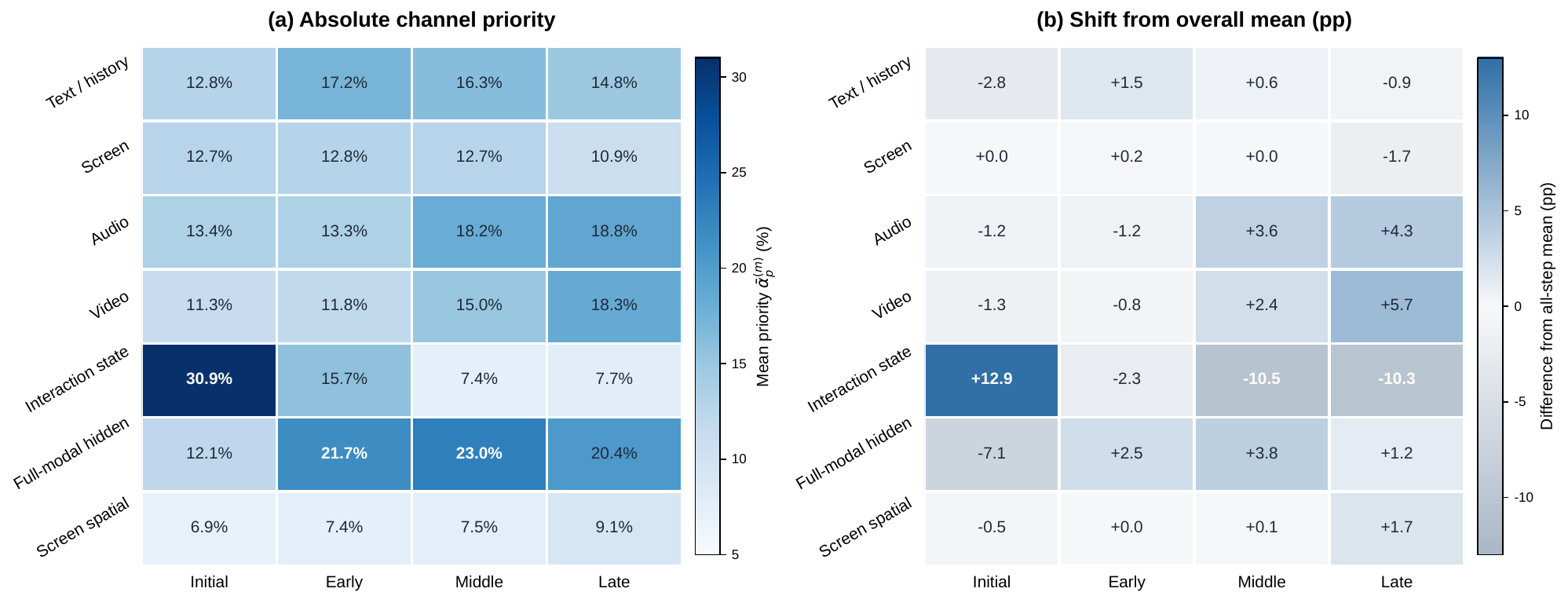}
    \caption{\textbf{Learned channel priorities reorder across interaction phases.}
(a) Mean priority $\bar{\alpha}^{(m)}$ per channel and phase on Qwen2.5-Omni; uniform allocation would be 14.3\%. (b) Deviation of each cell from that channel's all-step mean. FocusGUI trains on action annotations alone, with no
modality-relevance supervision.}
    \label{fig:heatmap}
    \vspace{-10pt}
\end{wrapfigure}

\begin{figure*}[t]
\centering
\includegraphics[width=1.0\linewidth]{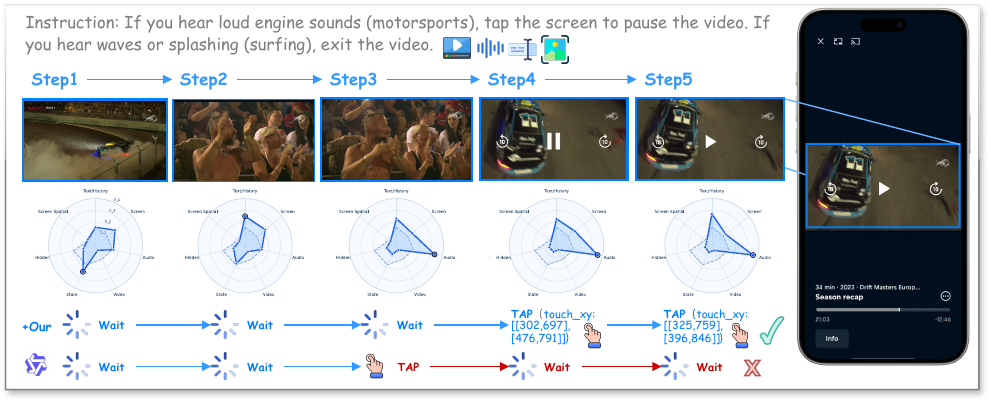}
\caption{\textbf{Case study: a decision that never appears on the screen.} The instruction conditions the action on what is heard, so the screen looks the same at every step while the deciding evidence sits in audio. FocusGUI waits and taps once the engine sounds arrive; the frozen backbone acts early on visual salience and cannot recover, since its own action re-enters the context of every later step. Radar plots show the channel priorities
$\alpha_t^{(m)}$ per step.}
\label{fig:exp6}
\end{figure*}

\textbf{Learned priorities reorder across an episode.}
Figure~\ref{fig:heatmap} averages $\alpha_t^{(m)}$ over test steps grouped by
interaction phase. Interaction state receives 30.9\% of the priority mass at
the opening step, more than twice a uniform share, and falls to 7.7\% by the
closing phase, moving from the highest-priority channel to the lowest; audio
and video move the other way, rising from 13.4\% and 11.3\% to 18.8\% and
18.3\%. No fixed weighting reproduces both ends of that range; only a
representation that reweights its inputs as the decision changes can, which is
exactly the step-adaptive behavior Focused Perception requires. The screen is never dominant: it
stays between 10.9\% and 12.8\%, below uniform at every phase, and the spatial
screen channel never exceeds 9.1\%. Prior work refines exactly these two
channels, while the channels that gain priority late are the streaming ones it
cannot reach. None of this is supervised: the ordering is
recovered from action annotations alone, and Fig.~\ref{fig:exp6} traces the
same shift within a single episode.

\begin{wraptable}{r}{0.6\textwidth}   
\vspace{-\intextsep}                   
\footnotesize
\setlength{\tabcolsep}{3pt}
\caption{Sensitivity to the boundary window $K$ and the compact focus dimension $d_z$ on Qwen2.5-Omni. The default ($K=4$, $d_z=64$, shaded) is fixed across all backbones and never tuned per benchmark.}
\label{tab:capacity_sweep}
\centering
\resizebox{\linewidth}{!}{%
\begin{tabular}{lcccc}
\toprule
& Overall TM & Overall EM & Overall GP & Temporal GP \\
\midrule
\multicolumn{5}{l}{\emph{Boundary window $K$ \; ($d_z=64$)}} \\
$K=1$    & 72.83 & 39.76 & 15.13 & 17.66 \\
\rowcolor{lightblue}
$K=4$    & 71.65 & 40.16 & 17.49 & 26.35 \\
$K=8$    & 70.28 & 37.80 & 15.72 & 16.04 \\
$K=16$   & 71.65 & 39.76 & 18.09 & 25.10 \\
\midrule
\multicolumn{5}{l}{\emph{Action-sufficient dimension $d_z$ \; ($K=4$)}} \\
$d_z=32$  & 70.47 & 33.27 & 10.95 & 12.45 \\
\rowcolor{lightblue}
$d_z=64$  & 71.65 & 40.16 & 17.49 & 26.35 \\
$d_z=128$ & 72.83 & 37.99 & 15.05 & 15.73 \\
$d_z=256$ & 71.23 & 38.19 & 18.28 & 15.83 \\
\bottomrule
\end{tabular}}
\end{wraptable}

\paragraph{Capacity sensitivity.} Table~\ref{tab:capacity_sweep} varies each
capacity parameter while holding the other at its default. Across a sixteenfold
range of $K$, Overall TM spans 2.6 points, EM 2.4, and GP 3.0, with no trend in
any of them. This matches the role $K$ plays in the design: $R_t$ is a
deterministic function of $z_t$, so a wider boundary window redistributes the
same compressed representation over more positions without admitting more of
the observation. Capacity is set by $d_z$ instead, and the sweep locates where
it binds. At $d_z=32$ the latent is too small to carry the action, costing 6.9
EM points and 6.5 GP points against the default. From 64 upward the method is
stable, with Overall TM spanning 1.6 points and Exact Match highest at the
default. FocusGUI requires no per-backbone or per-benchmark search
over either parameter, and we fix $K=4$ and $d_z=64$ throughout. For more comprehensive evaluation results, including performance across interaction depths on other backbones and capacity sensitivity analyses, please refer to the Appendix~\ref{app:exp}.

\section{Related Work}

\subsection{GUI Agents in Omni-Modal Environments}

GUI agents interact with digital environments by repeatedly translating user
intentions and interface observations into executable actions. Research across
mobile, web, and desktop platforms has established this perception--action
loop and progressively extended it from isolated grounding decisions to
long-horizon interaction~\citep{rawles2023androidinthewild,
deng2023mind2web,xie2024osworld,cheng2024seeclick,hong2024cogagent,zhang2025dualtapdualtaskadversarialprotector,chen2026skillharness}. Most
agents nevertheless organize their observation around the current screenshot,
with previous actions or screens retained as auxiliary context. Temporal GUI
benchmarks broaden this view by requiring agents to understand dynamic visual
content~\citep{jang2025videowebarena,zhang2026synchaininducingcomputeruseagent,chen2025gui}, while OmniGUI further
places each action within an interleaved observation of screen, video, audio,
and interaction history~\citep{henry2026omniguibenchmarkingguiagents}. These
developments shift GUI interaction from processing a sequence of interfaces to
acting over an evolving omni-modal state.

In parallel, omni-modal foundation models increasingly support unified or
streaming perception of text, vision, and audio~\citep{
xu2025qwen25omnitechnicalreport,fu2025vita15gpt4olevelrealtime,
cui2026minicpmo45realtimefullduplex}. Broader evaluations of real-time
assistants similarly emphasize online processing and continuity as multimodal
events unfold~\citep{lu2026omniinteract}. Such capabilities determine what a model can receive; how an agent should organize that evidence for the next GUI action remains open. The closest observation-side approach augments
computer-use agents with gated keyframes, audio transcription, and persistent
visual narration~\citep{li2026agentcomputer}. FocusGUI addresses the
complementary problem inside an omni-modal agent: given the complete
observation, it learns which evidence should shape the current action and how
to deliver that evidence to a frozen decoder.

\subsection{Perception and Grounding for GUI Agents}

Perception is central to GUI agents because an instruction must ultimately be
connected to an actionable interface element. One line of work strengthens
this connection through GUI-specific representation learning and large-scale
grounding data~\citep{cheng2024seeclick,hong2024cogagent,
ICLR2025_4ca0e369,wu2025iclr-osatlas}; another exposes explicit interface
structure to general-purpose models through screen parsing~\citep{
lu2024omniparserpurevisionbased, liu2026forwardbeatstwoinnerzoom, fu2026guilenscoarsetofinecroppinggui}. Recent visual agents also reduce redundant
screen tokens while jointly learning grounding and action generation~\citep{
Lin_2025_CVPR, liu2026zoomessencetrainlessgui}. Together, these methods improve how an agent understands the
current interface and determines where an action should be executed.

A complementary direction makes perception selective. Rather than processing
the full screenshot uniformly, visual search and test-time refinement narrow
the observation to promising regions~\citep{li2025screenspotpro,
Luo_2025_ICCV,liu2026continual}; active perception methods further decide when and how to crop, zoom, shift, or expand the visual field~\citep{chen2026guieyes,
Liu_2026_CVPR}. These approaches demonstrate the value of suppressing
irrelevant interface content, but their primary control axis remains spatial:
they determine which part of the current screen deserves closer inspection~\citep{zhang2026bamitrainingfreebiasmitigation}.
Omni-modal interaction introduces an additional axis, since relevance also
changes across information sources and action steps. FocusGUI extends focused perception from spatial localization to step-dependent representation construction over the complete observation, complementing existing GUI grounding mechanisms.

\section{Conclusion}
We identify Sequential Distraction as a primary challenge in multi-step omni-modal GUI interaction, where unfiltered observations can lead to compounding errors due to transient and sparse relevance. To address this, we formulate the concept of Focused Perception and implement it through FocusGUI, a lightweight adapter that dynamically prioritizes perceptual channels, compresses them into an action-sufficient latent, and delivers the result via a zero-initialized residual at the action boundary. This method relies solely on action annotations and leaves the underlying backbone completely frozen. Evaluations across four backbones demonstrate that FocusGUI substantially improves Type Match over existing enhancement methods and maintains high accuracy at deeper interaction steps where historical context typically degrades. While our current prioritization depends on the present observation, future work could extend this mechanism to anticipate channel relevance, which may help mitigate trajectory drift earlier in the interaction sequence.

\bibliography{references}
\bibliographystyle{preprint}

\clearpage
\newpage
\appendix

\etocdepthtag.toc{appendix}
\begingroup
\hypersetup{linkcolor=myblue}
\etocsettagdepth{mainmatter}{none}
\etocsettagdepth{appendix}{subsection}
\etocsettocstyle{\section*{Appendix Contents}}{}
\tableofcontents
\endgroup
\clearpage

\section{Notation}
\label{app:notation}
Table~\ref{tab:notation} lists the main symbols. Subscript $t$ denotes the
interaction step and superscript $(m)$ a perceptual channel.
\begin{table}[h]
\centering
\small
\setlength{\tabcolsep}{4pt}
\caption{Notation.}
\label{tab:notation}
\begin{tabular}{@{}llll@{}}
\toprule
Symbol & Meaning & Symbol & Meaning \\
\midrule
$t$ & Interaction step index & $P_m$ & Channel projector \\
$\mathcal{O}_t$ & Omni-modal observation & $e^{(m)}_t\!\in\!\mathbb{R}^{d}$ & Channel embedding \\
$g$ & User goal & $d$ & Shared channel dim.\ (128) \\
$I_t$ & Execution screenshot & $e^{(\mathrm{all})}_t$ & Integrated observation state \\
$V_t$ & Video observation & $q_t$ & Decision reference \\
$A_t$ & Audio observation & $r_m$ & Channel-identity embedding \\
$H_t$ & Interaction history & $s^{(m)}_t$ & Channel score \\
$S_t$ & Device state & $\tau$ & Softmax temperature (1.0) \\
$y_t$ & Ground-truth action tokens & $\alpha^{(m)}_t$ & Step-specific priority \\
$L_t$ & Number of action tokens & $c_t$ & Prioritized representation \\
$\mathcal{F}_{\Theta_0}$ & Frozen omni-modal backbone & $z_t\!\in\!\mathbb{R}^{d_z}$ & Compact focus latent \\
$\Theta_0$ & Frozen backbone parameters & $d_z$ & Action-sufficient dim.\ (64) \\
$\phi$ & Trainable FocusGUI parameters & $R_t$ & Residual, $\mathbb{R}^{K\times d_{\text{model}}}$ \\
$E_t$ & Prompt embeddings & $K$ & Residual slots (4) \\
$\widetilde{E}_t$ & Embeddings after injection & $J_t$ & $\{b_t-K,\dots,b_t-1\}$ \\
$d_{\text{model}}$ & Decoder hidden size & $\mathcal{S}_{J_t}(\cdot)$ & Placement at $J_t$, else zero \\
$b_t$ & Action boundary position & $\gamma$ & Residual scale (1.0) \\
$\mathcal{M}$ & Channel index set & $\mathcal{L}_{\mathrm{act}}$ & Action-token NLL, Eq.~(4) \\
$x^{(m)}_t$ & Raw channel feature & TM / EM / GP & Evaluation metrics \\
\bottomrule
\end{tabular}
\end{table}

\section{Discussion and Limitations}
FocusGUI is designed as a lightweight mechanism for improving how frozen omni-modal GUI agents organize step-dependent evidence, and it is complementary to advances in perception, grounding, and planning. Our current formulation computes channel priorities from the observation available at each interaction step, which keeps the method simple and model-agnostic; extending this mechanism to incorporate anticipatory signals about which evidence may become important in subsequent steps is a natural direction for future work. Our experiments span four omni-modal backbones and five interaction categories; extending focused perception to broader computer-use settings is a natural next step.

\section{More Experimental Results}
\label{app:exp}

\subsection{Capacity surface.} 
\begin{figure*}[ht]
\centering
\includegraphics[width=1.0\linewidth]{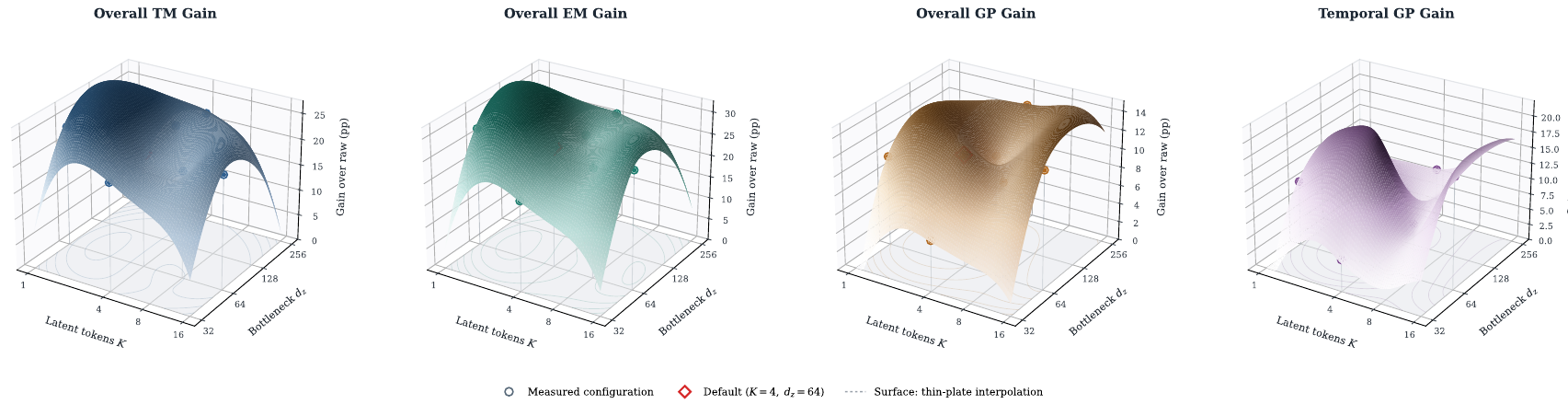}
\caption{\textbf{The capacity grid is flat around the default.} Overall TM, EM, GP, and Temporal GP as a function of the boundary window $K$ and the action-sufficient dimension $d_z$ on Qwen2.5-Omni. Circles mark measured configurations, and the surface is a thin-plate interpolation between them; the diamond marks the default $K=4$, $d_z=64$ used throughout the paper. The interpolated regions carry no measurement and are shown only to convey the shape of the grid.}
\label{fig:capacity_surface}
\end{figure*}

 Figure~\ref{fig:capacity_surface}
shows the full grid, with markers at measured configurations and a thin-plate
interpolation between them. For Overall TM, EM, and GP the surface is flat over
most of its extent: no ridge appears that a joint search over $K$ and $d_z$
would locate, and the default sits inside the flat region rather than on a peak
that depends on either parameter. Reading the markers rather than the
interpolated surface, the only configuration that separates from the rest is
$d_z=32$, where compression removes information the action needs.

\subsection{Interaction depth across backbones.} 

\begin{figure*}[t]
\centering
\includegraphics[width=1.0\linewidth]{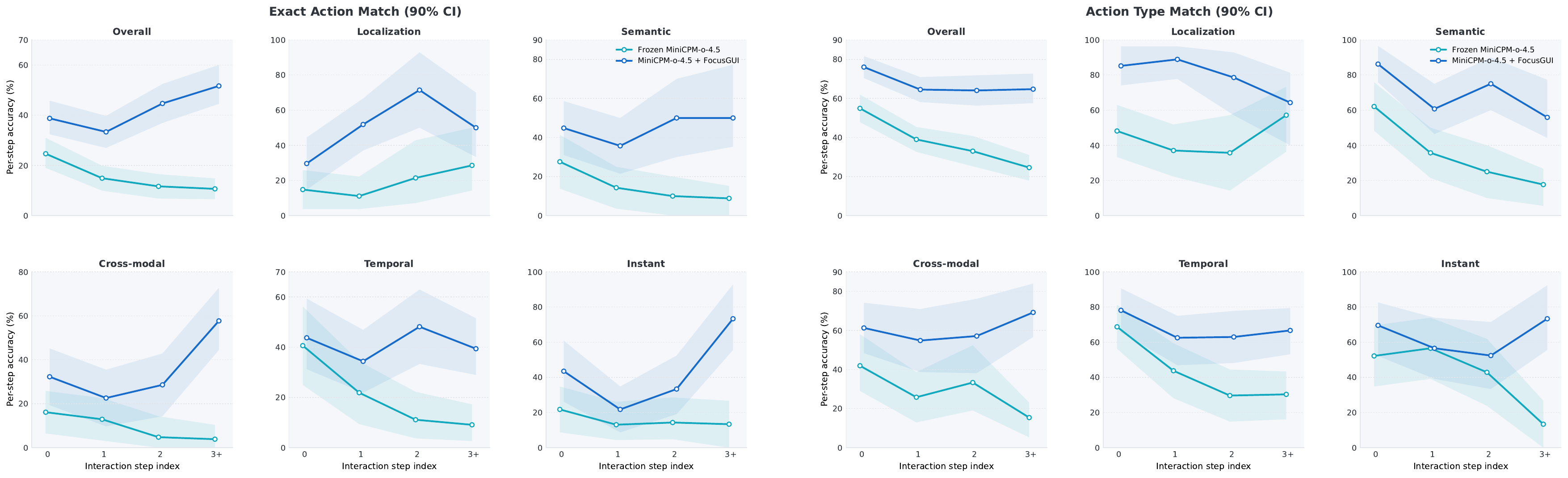}
\caption{\textbf{MiniCPM-o 4.5 across interaction depth.} Per-step Exact Match
(left) and Type Match (right) with steps binned by their index within the
episode and $3{+}$ collecting the fourth step onward. Shaded regions are $90\%$
episode-clustered bootstrap intervals. As on Qwen2.5-Omni, the frozen backbone declines with depth while FocusGUI holds, so the margin widens rather than staying constant.}
\label{fig:depth_minicpm}
\end{figure*}

\begin{figure*}[t]
\centering
\includegraphics[width=1.0\linewidth]{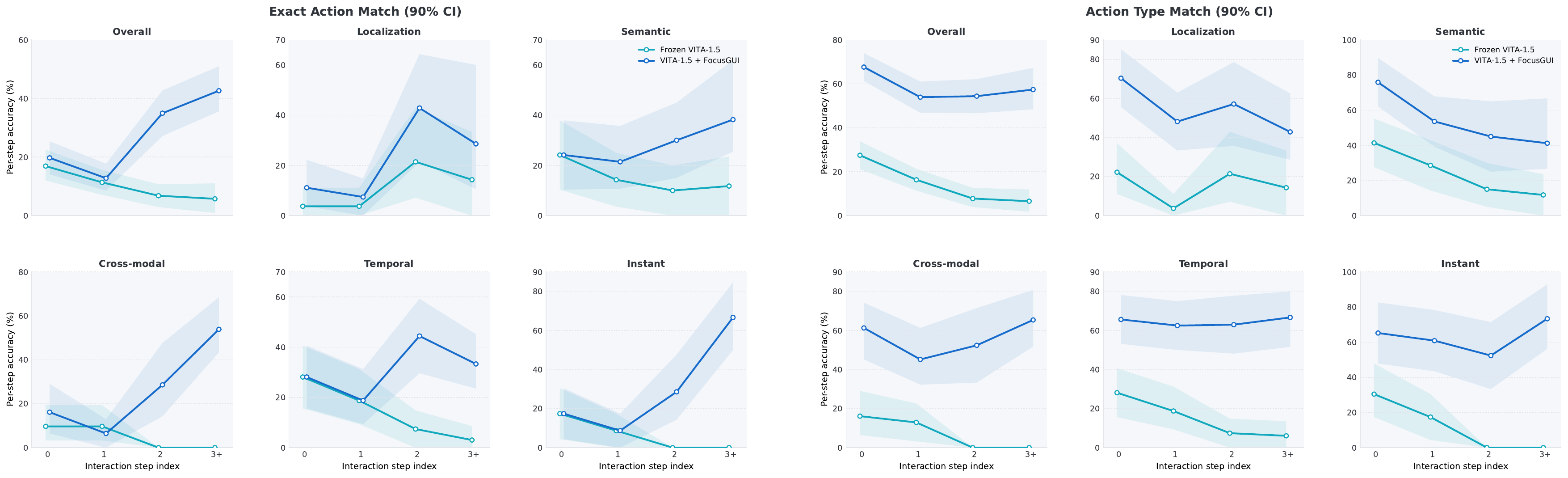}
\caption{\textbf{VITA-1.5 across interaction depth.} Per-step Exact Match
(left) and Type Match (right). VITA-1.5 is the weakest of the four backbones at
13.98 Overall TM, and the depth-dependent separation appears at that level as
well.}
\label{fig:depth_vita}
\end{figure*}

\begin{figure*}[t]
\centering
\includegraphics[width=1.0\linewidth]{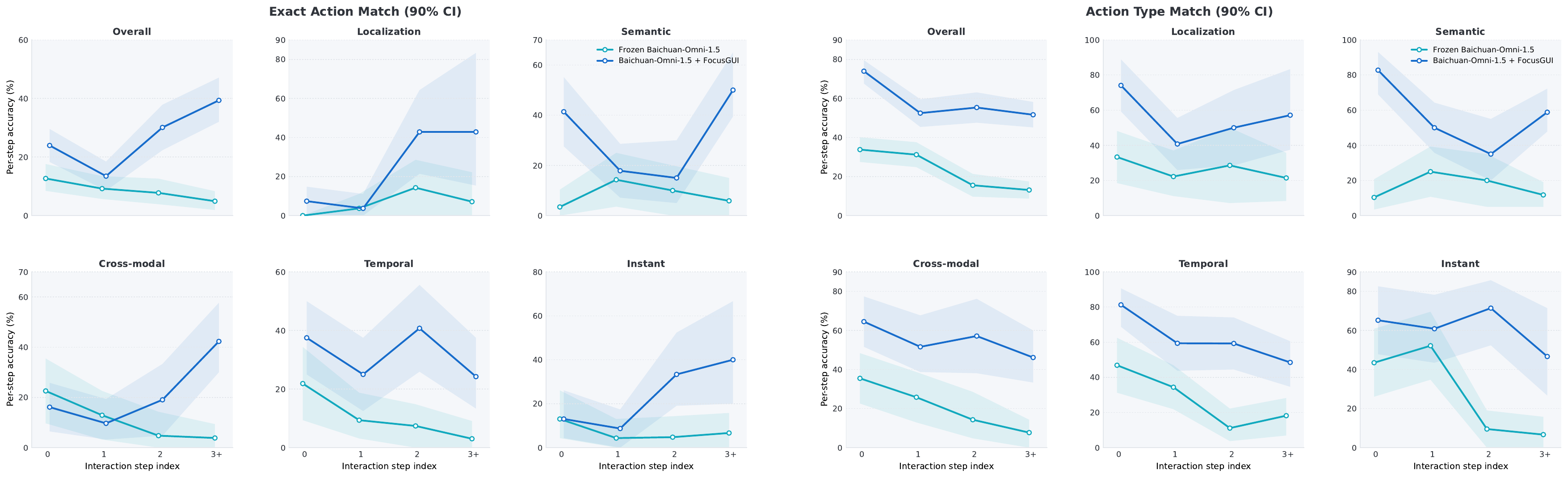}
\caption{\textbf{Baichuan-Omni-1.5 across interaction depth.} Per-step Exact
Match (left) and Type Match (right). The same widening margin holds on a third
backbone family, indicating that the decay is a property of the interaction
regime rather than of a particular model.}
\label{fig:depth_baichuan}
\end{figure*}

Figures~\ref{fig:depth_minicpm}
to~\ref{fig:depth_baichuan} repeat the analysis on MiniCPM-o 4.5, VITA-1.5,
and Baichuan-Omni-1.5, which span 13.98 to 36.02 Overall TM under frozen
inference and differ in architecture and multimodal representation. Each
frozen backbone declines as the episode advances, and each FocusGUI
counterpart holds its accuracy over the same range, so the margin widens with
depth in every case. On exact match the FocusGUI curve is flat or rising at the
depths where the frozen curve falls. The decay is therefore a property of the
interaction regime rather than of a particular model, and the mechanism that
addresses it transfers across backbone families with no change to the
architecture or to the training recipe.

\begin{table}[t]
\centering
\small
\setlength{\tabcolsep}{7pt}
\renewcommand{\arraystretch}{1.12}
\caption{Parameter-matched control on Qwen2.5-Omni. The concatenation variant
joins the perceptual channels with a fixed layout instead of scoring them
against a decision reference, and is sized to the same trainable budget. Both
configurations were trained under identical settings and evaluated on the same
508 test steps.}
\label{tab:param_matched}
\begin{tabular}{@{} l r S[table-format=2.2] S[table-format=2.2]
                      S[table-format=2.2] @{}}
\toprule[1pt]
 & & \multicolumn{3}{c}{\textcolor{steel}{\textbf{Metric (\%)}}} \\
\cmidrule(lr){3-5}
\textbf{Configuration} & \textbf{Params} & {TM} & {EM} & {GP} \\
\midrule
Parameter-matched concatenation & 1{,}444{,}913 & 71.85 & 33.46 & 10.14 \\
Dynamic prioritization          & 1{,}444{,}992 &
  \textbf{\textcolor{steel}{72.83}} & \textbf{\textcolor{steel}{37.01}} &
  \textbf{\textcolor{steel}{13.58}} \\
\midrule[0.3pt]
{\footnotesize $\Delta$\; (dyn.\ $-$ concat.)} & {\footnotesize $+79$} &
  {\footnotesize \gpos{0.98}} & {\footnotesize \gpos{3.54}} &
  {\footnotesize \gpos{3.44}} \\
\bottomrule[1pt]
\end{tabular}
\end{table}

\subsection{Prioritization, not capacity.} 
FocusGUI adds trainable parameters to a
frozen backbone, so part of its gain could in principle come from that budget
alone. Table~\ref{tab:param_matched} holds the budget fixed to within 79
parameters and replaces the decision reference with a fixed concatenation layout,
leaving the channels available but no longer scored per step. This control
differs from the uniform-priority variant in Table~\ref{tab:ablation_prioritization}:
uniform priorities average the channel embeddings with equal weights, so channel
identity is lost in the pooled representation, whereas concatenation keeps each
channel in its own slot and lets the downstream projection learn a fixed,
step-independent weighting. It is therefore the stronger static baseline, and
isolates per-step scoring as the only difference. Dynamic
prioritization improves all three metrics, with the largest gains on Exact
Match (+3.54) and Goal Progress (+3.44) and a smaller change in Type Match
(+0.98). The gain therefore appears where the mechanism is meant to act, on
executable grounding rather than on primitive selection, and it is not
explained by latent capacity.

\subsection{Paired Error Correction}
\begin{figure*}
\centering
\includegraphics[width=1.0\linewidth]{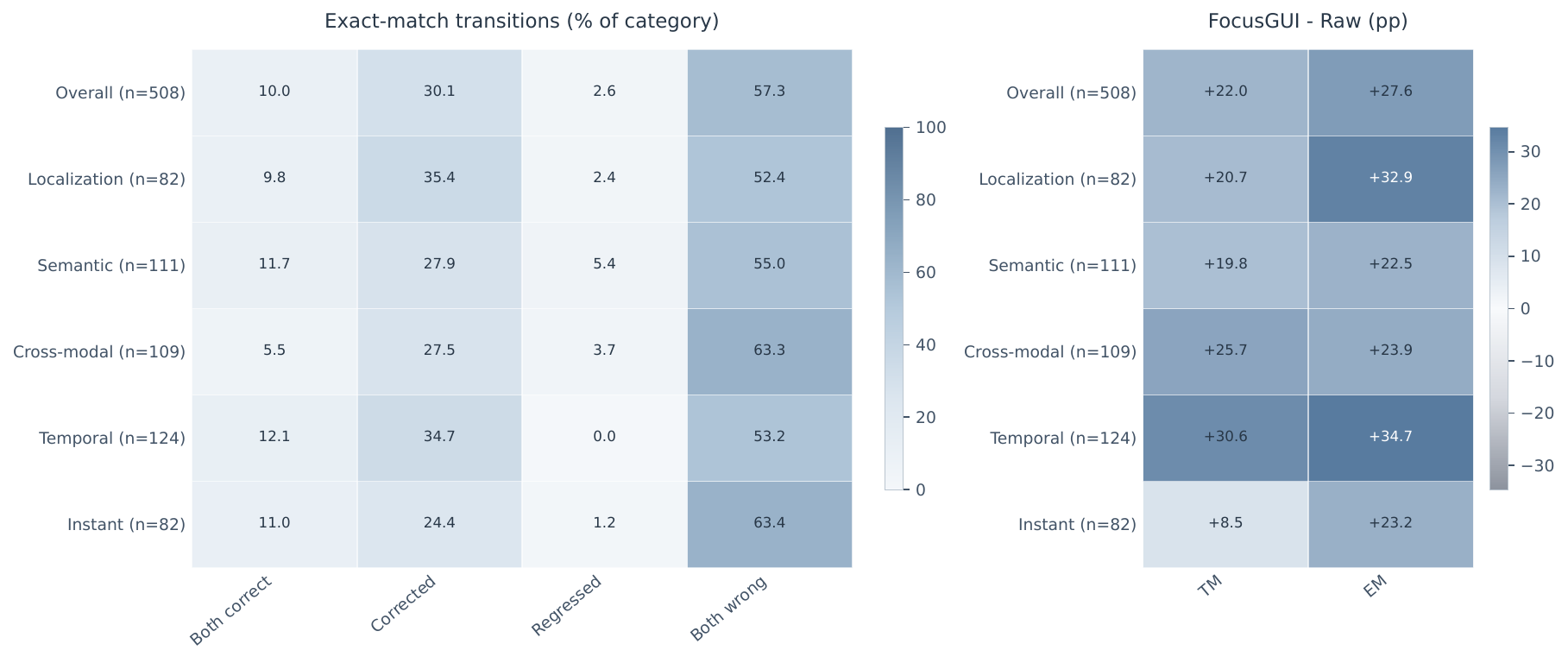}
\caption{Paired prediction outcomes across task categories. Left: proportions of steps that both models predict correctly, FocusGUI corrects, FocusGUI regresses, or both predict incorrectly. Right: improvements in type match (TM) and exact match (EM), measured in percentage points.}
\label{fig:paired_corrections}
\end{figure*}

Figure~\ref{fig:paired_corrections} examines how FocusGUI changes action correctness relative to the backbone across task categories. Corrections consistently outnumber regressions in all five categories, with 153 corrected predictions versus 13 regressions overall. The largest net improvement occurs in temporal reasoning, indicating that FocusGUI is particularly effective on tasks requiring decisions over evolving observations.

\section{More Implementation Details}
\label{app:impl}
\subsection{Prompt and action space.}
All four backbones and all five methods share the system prompt below, so the task description, the action vocabulary, and
the output format stay fixed while only the organization of the observation
differs. The action space follows OmniGUI and contains thirteen primitives:
three pointer actions carrying one coordinate, four swipes carrying a start and
an end coordinate, text entry carrying a string, four navigation and
termination actions without arguments, and \aname{NONE} for steps at which the
agent should observe rather than act. Coordinates are normalized to $[0,1000]$
and emitted as text through the frozen language-model head, which is what lets
FocusGUI retain the base model's action vocabulary instead of discretizing the
screen into a classification grid. The model returns one JSON object per step,
from which Type Match reads the primitive and Exact Match reads the arguments.
The residual window $\mathcal{J}_t$ occupies the four embedding positions
immediately preceding that object, so the injected latent is available to every
generated action token while the prompt itself is left unchanged.

\begin{promptbox}{System Prompt --- Mobile GUI Agent}
\footnotesize
\ptag{Role.}\; You are an Android GUI agent. Given a task, previous action
history, and the available screenshots, audio, and video, predict the next
executable action required to complete the task.

\smallskip
\ptag{Output.}\; Return \emph{exactly one} valid JSON object without
explanations, markdown, or additional text.

\smallskip
\ptag{Action space.}

\smallskip
{\setlength{\tabcolsep}{4.5pt}\renewcommand{\arraystretch}{1.18}
\begin{tabular}{@{} l l l @{}}
\toprule[0.6pt]
\multicolumn{1}{@{}l}{\textcolor{faint}{\scriptsize\textbf{ID}}} &
\multicolumn{1}{l}{\textcolor{faint}{\scriptsize\textbf{Action}}} &
\multicolumn{1}{l@{}}{\textcolor{faint}{\scriptsize\textbf{Arguments}}} \\
\midrule[0.3pt]
\aid{-1} & \aname{NONE}        & --- observation or waiting \\
\aid{0}  & \aname{TAP}         & \aname{coordinate:\,[x,\,y]} \\
\aid{1}  & \aname{DOUBLE\_TAP} & \aname{coordinate:\,[x,\,y]} \\
\aid{2}  & \aname{LONG\_PRESS} & \aname{coordinate:\,[x,\,y]} \\
\aid{3}  & \aname{SWIPE\_UP}    & \aname{start\_coordinate,\;end\_coordinate} \\
\aid{4}  & \aname{SWIPE\_DOWN}  & \aname{start\_coordinate,\;end\_coordinate} \\
\aid{5}  & \aname{SWIPE\_LEFT}  & \aname{start\_coordinate,\;end\_coordinate} \\
\aid{6}  & \aname{SWIPE\_RIGHT} & \aname{start\_coordinate,\;end\_coordinate} \\
\aid{7}  & \aname{INPUT}       & \aname{text:\,string} \\
\aid{8}  & \aname{BACK}        & --- \\
\aid{9}  & \aname{HOME}        & --- \\
\aid{10} & \aname{TASK\_COMPLETE}   & --- \\
\aid{11} & \aname{TASK\_IMPOSSIBLE} & --- \\
\bottomrule[0.6pt]
\end{tabular}}

\smallskip
\ptag{Conventions.}
\begin{itemize}[leftmargin=1.1em,itemsep=0.5pt,topsep=1pt,parsep=0pt]
  \item Coordinates use $[x, y]$ order, normalized to $[0, 1000]$; the current
        screenshot is the execution surface.
  \item Return \aname{NONE} when observation or waiting is required;
        \aname{TASK\_COMPLETE} when the goal has already been achieved;
        \aname{TASK\_IMPOSSIBLE} only when the task cannot be completed.
\end{itemize}
\end{promptbox}

\subsection{Training target and inference.}
During training, the ground-truth action is serialized as a compact JSON
string and appended only as the assistant target, e.g.,
\texttt{\{"action\_type":0,"coordinate":[421,736]\}}.
Loss is computed exclusively over assistant action tokens; all system and
user tokens are masked. During inference, the target is absent and the
original frozen LM head autoregressively generates the action using greedy
decoding with beam size one.

The raw backbone and FocusGUI receive exactly the same semantic prompt,
multimodal observations, action history, action schema, and decoding
configuration. FocusGUI does not append textual instructions or modality hints.
Its selected evidence is projected into continuous latent residuals at the
assistant action boundary before generation by the original decoder and
LM head.

\subsection{Step template.}
Each step is presented with the template below. The
multimodal blocks appear in a fixed order, with the history screenshot first
when one exists, then the temporal and acoustic observations, and the execution
screen last, so the screen the action applies to is always the most recent
visual input. The goal and the step index follow, then the actions already
executed in this episode, rendered as text rather than as structured state. The
observation elements therefore reach the backbone exactly as they would without
FocusGUI, and the channels of Sec.~\ref{sec:method} are read from this same
input rather than replacing any part of it. The closing line ends the prompt and
fixes the action boundary $b_t$ at which the residual is applied.

\begin{promptbox}{User Prompt --- Step Template}
\footnotesize

\newcommand{\mtag}[1]{\colorbox{chipbg}{\textcolor{faint}{\texttt{#1}}}}
\newcommand{\ph}[1]{\textcolor{steel}{\texttt{\{#1\}}}}

\begin{tabular}{@{} l l @{}}
\textcolor{faint}{\emph{Optional history}}     & \mtag{<HistoryScreenshot>} \\[2pt]
\textcolor{faint}{\emph{Temporal observation}} & \mtag{<Video\_t>} \\[2pt]
\textcolor{faint}{\emph{Acoustic observation}} & \mtag{<Audio\_t>} \\[2pt]
\textcolor{faint}{\emph{Execution screen}}     & \mtag{<Screen\_t>} \\
\end{tabular}

\smallskip
\ptag{Goal.}\; \ph{task\_instruction}
\quad
\ptag{Step.}\; \ph{step\_index}

\smallskip
\ptag{Action history.}

\smallskip
{\ttfamily\footnotesize
\begin{tabular}{@{} l l @{}}
Step 0:     & \ph{previous\_action\_0} \\
\ldots      & \\
Step $t-1$: & \ph{previous\_action\_t-1} \\
\end{tabular}}

\smallskip
Analyze the current observation and output the next action as JSON.
\end{promptbox}

\subsection{Experiment Details}
We construct episode-disjoint training, validation, and test sets containing
1,807, 257, and 508 interaction steps from 495, 71, and 142 trajectories,
respectively. All feature transformations and normalization statistics are
fitted using the training split only. Validation data are used exclusively
for checkpoint selection, and the test split is evaluated once after model
selection. At inference time, the original LM head generates one structured
action using greedy decoding with beam size one. Raw backbones and their FocusGUI variants use identical observations, prompts, action histories, action spaces, and decoding configurations.  On Qwen2.5-Omni, FocusGUI contains 1.445M trainable parameters compared with 8.932B frozen Thinker parameters, corresponding to only $0.0162\%$ trainable parameters. We train a separate FocusGUI adapter for each of the four backbones, with no adapter parameters shared across backbone families. Across the evaluated backbones, the adapter size ranges from 1.15M to 1.45M parameters. All experiments are conducted using two NVIDIA
L20 GPUs.

\begin{table}[t]
\centering
\small
\caption{Default FocusGUI implementation configuration.}
\begin{tabular}{lc}
\toprule
Configuration & Value \\
\midrule
Shared channel dimension $d$ & 128 \\
Action-sufficient dimension $d_z$ & 64 \\
Residual slots $K$ & 4 \\
Residual initialization & Zero \\
Residual scale $\gamma$ & 1.0 \\
Channel temperature $\tau$ & 1.0 \\
Dropout & 0.05 \\
Optimizer & AdamW \\
Learning rate & $2\times10^{-4}$ \\
Weight decay & $10^{-3}$ \\
Maximum epochs / patience & 3 / 2 \\
Gradient clipping & 1.0 \\
Decoding & Greedy, beam size 1 \\
Qwen FocusGUI parameters & 1.445M \\
Frozen Qwen Thinker parameters & 8.932B \\
Trainable proportion & 0.0162\% \\
\bottomrule
\end{tabular}
\end{table}

\begin{table}[t]
\centering
\small
\setlength{\tabcolsep}{8pt}
\renewcommand{\arraystretch}{1.15}
\caption{What the two configurations receive at inference. Every row above the
rule is shared, so the columns are indistinguishable until the last two entries.
$\mathcal{J}_t$ denotes the $K$ embedding positions preceding the action
boundary.}
\label{tab:controlled_inputs}
\begin{tabular}{@{} l c c @{}}
\toprule
& Frozen backbone & $+$ FocusGUI \\
\midrule
System prompt              & \multicolumn{2}{c}{identical} \\
Multimodal observations    & \multicolumn{2}{c}{identical} \\
Action history             & \multicolumn{2}{c}{identical} \\
Action schema              & \multicolumn{2}{c}{identical} \\
Decoding                   & \multicolumn{2}{c}{greedy, beam size 1} \\
Textual demonstrations     & \multicolumn{2}{c}{none} \\
Modality-relevance hints   & \multicolumn{2}{c}{none} \\
Backbone weights           & \multicolumn{2}{c}{frozen} \\
\midrule
Trainable parameters       & 0 & 1.15M--1.45M \\
Input at $\mathcal{J}_t$   & $E_{t,\mathcal{J}_t}$ & $E_{t,\mathcal{J}_t} + \gamma R_t$ \\
\bottomrule
\end{tabular}
\end{table}

\subsection{Controlled inputs.} 
Table~\ref{tab:controlled_inputs} lists what the
frozen backbone and FocusGUI receive at inference. The prompt, the observations,
the action history, the schema, and the decoding procedure are identical, and
neither configuration is given textual demonstrations or any hint about which
modality matters at a step. The two differ in the parameters that build $z_t$
and in the residual added at $\mathcal{J}_t$, and
Table~\ref{tab:param_matched} separates those by holding the parameter budget
fixed while removing the per-step scoring. A gain over the frozen backbone
therefore reflects the content of $R_t$ rather than a longer prompt, an
in-context example, or a different decoding rule.

\newpage
\section{Algorithm}
\label{app:algorithm}

Algorithm~\ref{alg:forward} details the per-step FocusGUI forward pass, which is
shared by training and inference: it exposes the perceptual channels, scores them
against the decision reference $q_t$, compresses the result into $z_t$, and adds
the zero-initialized residual at the $K$ positions preceding the action boundary.
Algorithm~\ref{alg:train} gives the training and inference procedures built on it.
\textcolor{myblue}{Blue} marks everything FocusGUI contributes; all remaining lines are the frozen
backbone's original pathway. Gradients flow through the frozen decoder to $\phi$,
but no backbone parameter is updated and the prompt itself is never modified.
 
\begin{algorithm}[t]
\caption{FocusGUI forward pass at interaction step $t$}
\label{alg:forward}
\begin{algorithmic}[1]
\Require Observation $\mathcal{O}_t$; frozen backbone $\mathcal{F}_{\Theta_0}$;
         trainable \textcolor{myblue}{$\phi=\{P_m,P_q,\{r_m\},W_e,W_q,v,P_s,W_p,b_p\}$};
         hyperparameters $K,d_z,\tau,\gamma$
\Ensure Modified embeddings \textcolor{myblue}{$\widetilde{E}_t$}; channel priorities \textcolor{myblue}{$\{\alpha^{(m)}_t\}$}
\State $E_t,\; \{x^{(m)}_t\}_{m\in\mathcal{M}},\; e^{(\mathrm{all})}_t \gets \mathcal{F}_{\Theta_0}.\textsc{Encode}(\mathcal{O}_t)$
       \Comment{prompt embeddings, channels, integrated state}
\ForAll{$m\in\mathcal{M}$} \Comment{\textcolor{myblue}{channel decomposition (CD)}}
    \State \textcolor{myblue}{$e^{(m)}_t \gets P_m\!\left(x^{(m)}_t\right)\in\mathbb{R}^{d}$}
\EndFor
\State \textcolor{myblue}{$q_t \gets P_q\!\left(e^{(\mathrm{all})}_t\right)$} \Comment{decision reference, recomputed every step}
\ForAll{$m\in\mathcal{M}$} \Comment{\textcolor{myblue}{step-adaptive prioritization (SP)}}
    \State \textcolor{myblue}{$s^{(m)}_t \gets v^{\top}\tanh\!\left(W_e\big(e^{(m)}_t+r_m\big)+W_q q_t\right)$}
\EndFor
\State \textcolor{myblue}{$\alpha^{(m)}_t \gets \dfrac{\exp\!\big(s^{(m)}_t/\tau\big)}{\sum_{j\in\mathcal{M}}\exp\!\big(s^{(j)}_t/\tau\big)}$,
       \quad $c_t \gets \sum_{m\in\mathcal{M}} \alpha^{(m)}_t e^{(m)}_t$}
\State \textcolor{myblue}{$z_t \gets P_s(c_t)\in\mathbb{R}^{d_z}$} \Comment{\textcolor{myblue}{action-sufficient projection (AP)}, $d_z<d$}
\State \textcolor{myblue}{$R_t \gets \textsc{Reshape}\!\left(W_p z_t+b_p\right)\in\mathbb{R}^{K\times d_{\text{model}}}$}
       \Comment{$W_p,b_p$ zero-initialized $\Rightarrow R_t=0$ at the start of training}
\State $J_t \gets \{b_t-K,\dots,b_t-1\}$ \Comment{$K$ prompt positions before the action boundary}
\State $\widetilde{E}_t \gets E_t + \textcolor{myblue}{\mathcal{S}_{J_t}(\gamma R_t)}$
       \Comment{\textcolor{myblue}{additive residual alignment (RA)}; $\mathcal{O}_t$ is neither pruned nor extended}
\State \Return $\widetilde{E}_t,\;\{\alpha^{(m)}_t\}_{m\in\mathcal{M}}$
\end{algorithmic}
\end{algorithm}
 
\begin{algorithm}[t]
\caption{Training from action supervision, and inference}
\label{alg:train}
\begin{algorithmic}[1]
\Procedure{Train}{$\mathcal{D}_{\mathrm{tr}},\mathcal{D}_{\mathrm{val}}$}
    \State Initialize $\phi$; set \textcolor{myblue}{$W_p\gets 0$, $b_p\gets 0$}; freeze encoder, decoder, and LM head ($\Theta_0$)
    \For{epoch $=1$ \textbf{to} $E_{\max}$}
        \For{mini-batch $\mathcal{B}\subset\mathcal{D}_{\mathrm{tr}}$}
            \For{step $t\in\mathcal{B}$}
                \State \textcolor{myblue}{$\widetilde{E}_t,\,\cdot \gets$ \Call{FocusGUIForward}{$\mathcal{O}_t$}}
                       \Comment{Algorithm~\ref{alg:forward}}
                \State $\ell_t \gets -\frac{1}{L_t}\sum_{k=1}^{L_t}\log p_{\Theta_0}\!\left(y_{t,k}\mid y_{t,<k},\textcolor{myblue}{\widetilde{E}_t}\right)$
                       \Comment{teacher forcing; loss on action tokens only}
            \EndFor
            \State $\mathcal{L}_{\mathrm{act}}(\phi)\gets \frac{1}{|\mathcal{B}|}\sum_{t\in\mathcal{B}}\ell_t$
            \State \textcolor{myblue}{$\phi \gets \textsc{AdamW}\!\left(\phi,\;\textsc{ClipGrad}\big(\nabla_\phi\mathcal{L}_{\mathrm{act}}\big)\right)$}
                   \Comment{\textcolor{myblue}{$\nabla_{\Theta_0}$ discarded; $\Theta_0$ unchanged}}
        \EndFor
        \State Evaluate on $\mathcal{D}_{\mathrm{val}}$; keep the best checkpoint; stop early if patience is exhausted
    \EndFor
    \State \Return \textcolor{myblue}{$\phi^\star$}
\EndProcedure
\Statex
\Procedure{Infer}{$g$}
    \State $H_0\gets\emptyset$
    \For{$t=0,1,2,\dots$ until termination or step budget}
        \State Observe $\mathcal{O}_t=\{g,I_t,V_t,A_t,H_t,S_t\}$
        \State \textcolor{myblue}{$\widetilde{E}_t,\,\{\alpha^{(m)}_t\} \gets$ \Call{FocusGUIForward}{$\mathcal{O}_t$}}
        \State $\hat{y}_t \gets \mathcal{F}_{\Theta_0}.\textsc{Decode}\!\left(\textcolor{myblue}{\widetilde{E}_t}\right)$
               \Comment{original LM head, greedy decoding, beam size 1}
        \State Execute $\hat{y}_t$; \; $H_{t+1}\gets H_t\cup\{\hat{y}_t\}$
               \Comment{the executed action re-enters the next observation}
    \EndFor
\EndProcedure
\end{algorithmic}
\end{algorithm}

\section{Case Study}
\label{app:case}
To demonstrate the comprehensiveness and robustness of FocusGUI across diverse interaction paradigms, we present two distinct case studies encompassing both cross-modal temporal reasoning and multi-step semantic localization. Real-world GUI operation requires agents to seamlessly transition between different cognitive modes: some tasks demand patient monitoring of streaming sensory inputs (e.g., waiting for an auditory cue), while others require precise, sequential spatial navigation through dense interfaces.

\begin{figure*}
    \centering
    \includegraphics[width=1.0\linewidth]{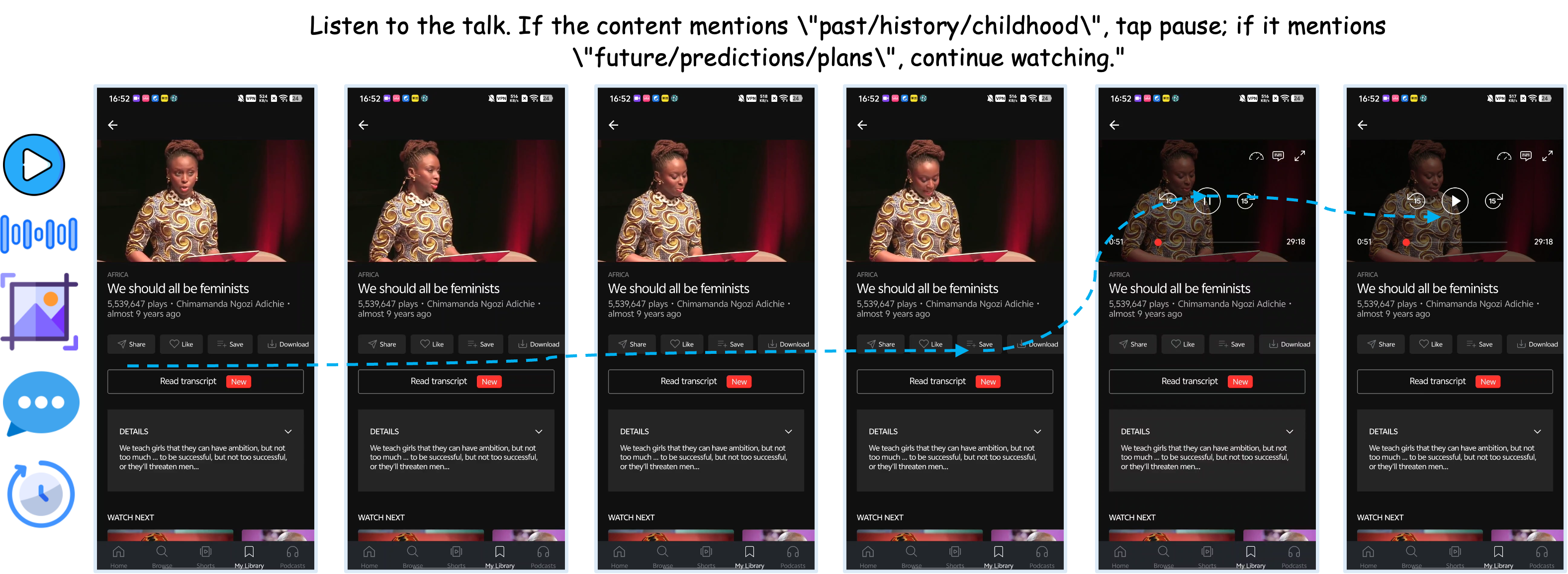}
    \caption{Case study on cross-modal temporal reasoning. The agent is instructed to pause a video based on specific auditory cues ("past/history/childhood"). The visual interface remains largely static, forcing the model to rely on streaming audio comprehension over multiple steps. FocusGUI successfully avoids premature visual actions, dynamically shifting priority to the audio channel and executing the spatial action (tapping pause) precisely when the semantic condition is met.}
    \label{fig:case1}
\end{figure*}

\begin{figure*}
    \centering
    \includegraphics[width=1.0\linewidth]{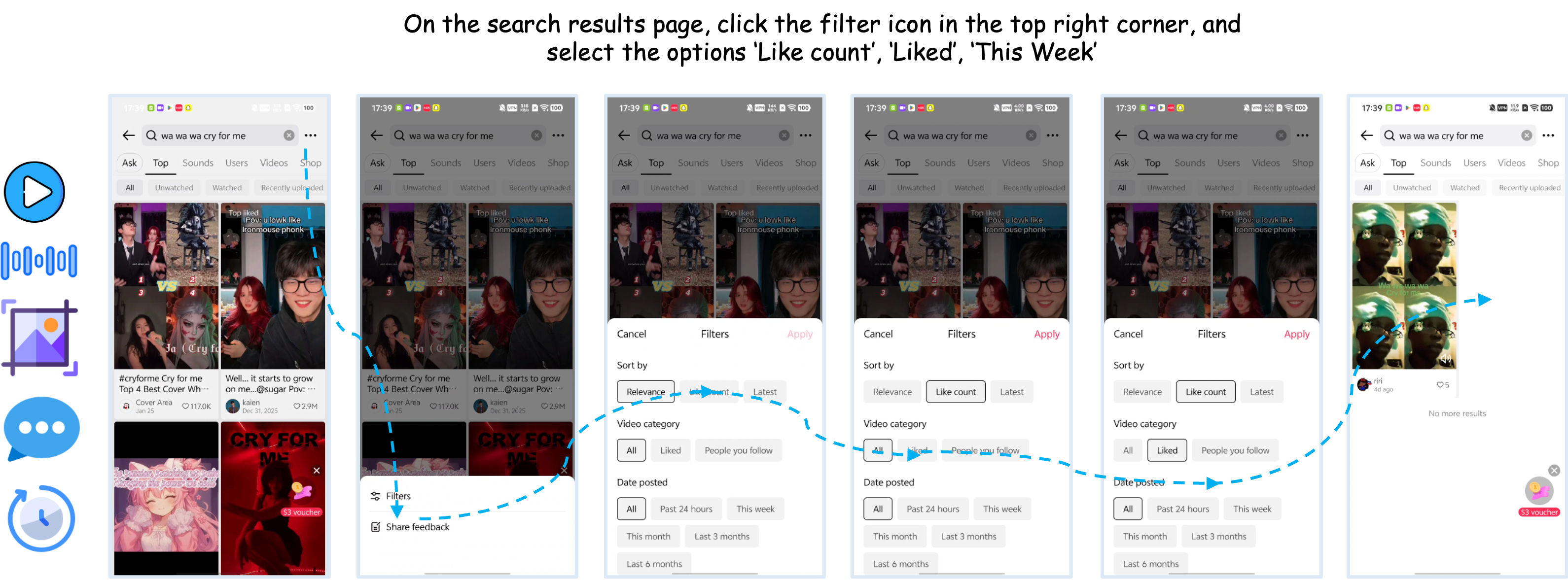}
    \caption{Case study on multi-step semantic planning and spatial localization. The agent executes a complex sequence of interface interactions based on a detailed text instruction. It successfully navigates through dynamic UI changes—locating a small filter icon, understanding the popup menu, and sequentially selecting "Like count", "Liked", and "This Week". This demonstrates FocusGUI's capacity for precise spatial grounding and robust state tracking in visually dense environments.}
    \label{fig:case2}
\end{figure*}

The following examples illustrate how FocusGUI successfully navigates these contrasting extremes by dynamically prioritizing relevant perceptual channels. Figure~\ref{fig:case1} highlights a scenario where the decisive signal is entirely auditory and temporal, requiring the agent to ignore static visual saliency until a specific semantic condition is met. Conversely, Figure~\ref{fig:case2} showcases a highly spatial and semantic task, where the agent must execute a precise sequence of UI interactions (opening filters, selecting specific nested options) based on a composite text instruction. Together, these cases confirm that FocusGUI's step-adaptive prioritization effectively generalizes across fundamentally different types of GUI challenges, preventing Sequential Distraction in both streaming and structurally complex environments.

\newpage
\section{Theoretical Analysis}
\label{app:theory}

We trace how the latent $z_t$ reaches the action logits through the frozen
decoder, and show that it acts through two pathways with distinct roles:
one changes \emph{what} the decoder reads from a position and suffices to
select the action primitive; the other changes \emph{where} it reads and is
the only route by which arguments can be corrected.

The decoder is $L$ frozen pre-LN blocks on a residual stream
$h^{(\ell)}_i \in \mathbb{R}^d$, with per-head attention
$q_k = W_Q\mathrm{LN}(h_k)$, $\kappa_i = W_K\mathrm{LN}(h_i)$,
$v_i = W_V\mathrm{LN}(h_i)$,
$a_{k\cdot}=\mathrm{softmax}(\langle q_k,\kappa_\cdot\rangle/\sqrt{d_h})$,
$o_k = W_O\sum_i a_{ki}v_i$, and logits $f_k = W_U\mathrm{LN}(h^{(L)}_k)$.
FocusGUI sets $h^{(0)}_i = e_i + \gamma R^{(i)}\mathbb{1}[i\in J_t]$ with
$R = \mathrm{Reshape}(W_p z_t + b_p)$ and $J_t=\{b_t-K,\dots,b_t-1\}$. Write
$\delta^{(\ell)}_i := \partial h^{(\ell)}_i/\partial\gamma$ and
$\Lambda_i$ for the LayerNorm Jacobian. Since the finite change
$\Delta f_k = \gamma\int_0^1 \partial_\gamma f_k|_{\gamma u}\,du$, the
decomposition below holds without linearization. Because
$\delta^{(0)}_{J_t}$ ranges over an affine set with direction space
$\mathrm{col}(W_p)$ and the frozen Jacobians act linearly, every
$\delta^{(\ell)}_i$ lies in a family of dimension at most $d_z$: the adapter
steers the frozen computation through a $d_z$-dimensional control, and what
the network does with it is fixed by $\Theta_0$.

For an action position $k \ge b_t$, differentiating $o_k$ gives
\begin{equation}
  \frac{\partial o_k}{\partial \gamma}
  = W_O\underbrace{\sum_i a_{ki} W_V \Lambda_i \delta_i}_{\textbf{(V) value}}
  + W_O\underbrace{\sum_i a_{ki}\,\dot s_{ki}\big(v_i - \bar v_k\big)}_{\textbf{(A) attention}},
  \qquad
  \bar v_k = \sum_j a_{kj} v_j,
  \label{eq:two-paths}
\end{equation}
where the second form follows from the softmax Jacobian and
$\sum_i \partial_\gamma a_{ki} = 0$, and
\begin{equation}
  \dot s_{ki}
  = \tfrac{1}{\sqrt{d_h}}\big(
      \underbrace{\langle \delta_k,\, \Lambda_k^\top W_Q^\top W_K\,\mathrm{LN}(h_i)\rangle}_{\text{query steering}}
    + \underbrace{\langle q_k,\, W_K \Lambda_i \delta_i \rangle}_{\text{key displacement}}\big).
  \label{eq:score-dot}
\end{equation}
Pathway (A) conserves attention mass and moves the read along the contrasts
$v_i - \bar v_k$: it adds no content and only reallocates which existing
context the action token consumes. Content enters through (V) alone, and only
at positions where $\delta \ne 0$.

\begin{proposition}[Depth-staged re-routing]
\label{prop:stages}
At layer $1$, $\delta_k = 0$ for all $k \ge b_t$, so only key displacement is
active and \eqref{eq:two-paths} reweights attention within $J_t$. From layer
$2$ on, the first layer's attention has written $\delta_k \ne 0$, and query
steering makes $\dot s_{ki} \ne 0$ for every unmasked $i$: a residual confined
to $K$ positions redistributes attention over the entire observation.
Moreover, by induction under the causal mask, $\delta^{(\ell)}_i = 0$ for all
$i < b_t - K$ at every layer, so the content $v_i$ at every evidence position
in the observation is exactly independent of $\gamma$.
\end{proposition}

\paragraph{Consequences for action generation.}
The two output families of a GUI action consume different pathways. The
\emph{primitive} is a categorical choice: flipping it from $b$ to $a$ only
requires $W_U\Lambda^{(L)}_k\delta^{(L)}_k$ to have a component along
$w_a - w_b$ exceeding the current logit gap, which pathway (V) supplies
directly. \emph{Arguments} (coordinates, typed text) are produced by
copy-type readout from evidence positions, which satisfy $i < b_t - K$; by
Proposition~\ref{prop:stages} their content is untouched, so argument
corrections arise only through pathway (A) re-routing mass onto the correct
evidence. This is why the injection must be \emph{additive}: replacing
$e_{J_t} + \gamma R$ by $R$ leaves query steering intact but overwrites the
$K$ most recent, most heavily attended context positions with $z_t$-derived
content of dimension at most $d_z$, destroying their value as evidence.

\end{document}